\documentclass[acmtog,nonacm,screen]{acmart}
\definecolor{GXYColor}{rgb}{1.0,0.6,0.0}
\definecolor{HXGColor}{rgb}{1.0,0.3,0.0}

\usepackage{subfigure}
\usepackage{amsfonts}
\usepackage{enumitem}

\AtBeginDocument{%
  }

\begin{document}

\title{Towards Realistic Example-based Modeling via 3D Gaussian Stitching}
% \emph{RealExam}: 

\author{Xinyu Gao}
\authornote{equal contribution}
\orcid{0009-0007-1079-6451}
\affiliation{%
  \institution{State Key Lab of CAD\&CG, Zhejiang University}
  \country{China}
}
\author{Ziyi Yang}
\authornotemark[1]
\email{14ziyiyang@gmail.com}
\orcid{0000-0002-9318-4704}
\affiliation{%
  \institution{State Key Lab of CAD\&CG, Zhejiang University}
  \country{China}
}
\author{Bingchen Gong}
\authornotemark[1]
\email{gongbingchen@gmail.com}
\orcid{0000-0001-6459-6972}
\affiliation{%
    \institution{The Chinese University of Hong Kong}
    \country{Hong Kong}
}
\author{Xiaoguang Han}
\email{hanxiaoguang@cuhk.edu.cn}
\orcid{0000-0003-0162-3296}
\affiliation{%
    \institution{The Chinese University of Hong Kong, Shenzhen}
    \country{China}
}
\author{Sipeng Yang}
\email{12121024@zju.edu.cn}
\orcid{0000-0002-8141-2335}
\affiliation{%
    \institution{State Key Lab of CAD and CG, Zhejiang University}
    \country{China}
}
\author{Xiaogang Jin}
\email{jin@cad.zju.edu.cn}
\orcid{0000-0001-7339-2920}
\affiliation{%
    \institution{State Key Lab of CAD and CG, Zhejiang University}
    \country{China}
}
% \author[\href{https://github.com/ingra14m/gs_stitching_website}{homepage}]{}

\renewcommand{\shortauthors}{Xinyu Gao, et al.}

\begin{abstract}
Using parts of existing models to rebuild new models, commonly termed as example-based modeling, is a classical methodology in the realm of computer graphics. Previous works mostly focus on shape composition, making them very hard to use for realistic composition of 3D objects captured from real-world scenes. This leads to combining multiple NeRFs into a single 3D scene to achieve seamless appearance blending. 
However, the current SeamlessNeRF method struggles to achieve interactive editing and harmonious stitching for real-world scenes due to its gradient-based strategy and grid-based representation.
To this end, we present an example-based modeling method that combines multiple Gaussian fields in a point-based representation using sample-guided synthesis. 
Specifically, as for composition, we create a GUI to segment and transform multiple fields in real time, easily obtaining a semantically meaningful composition of models represented by 3D Gaussian Splatting (3DGS). For texture blending, due to the discrete and irregular nature of 3DGS, straightforwardly applying gradient propagation as SeamlssNeRF is not supported. Thus, a novel sampling-based cloning method is proposed to harmonize the blending while preserving the original rich texture and content. Our workflow consists of three steps: 1) real-time segmentation and transformation of a Gaussian model using a well-tailored GUI, 2) KNN analysis to identify boundary points in the intersecting area between the source and target models, and 3) two-phase optimization of the target model using sampling-based cloning and gradient constraints. Extensive experimental results validate that our approach significantly outperforms previous works in terms of realistic synthesis, demonstrating its practicality. More demos are available at \href{https://ingra14m.github.io/gs_stitching_website}{homepage}.

\end{abstract}


% \ccsdesc[500]{Computing methodologies~Point-based models}
% \ccsdesc[300]{Computing methodologies~Volumetric models}
% \ccsdesc[100]{Computing methodologies~Computational photography}


\begin{teaserfigure}
  \includegraphics[width=\textwidth]{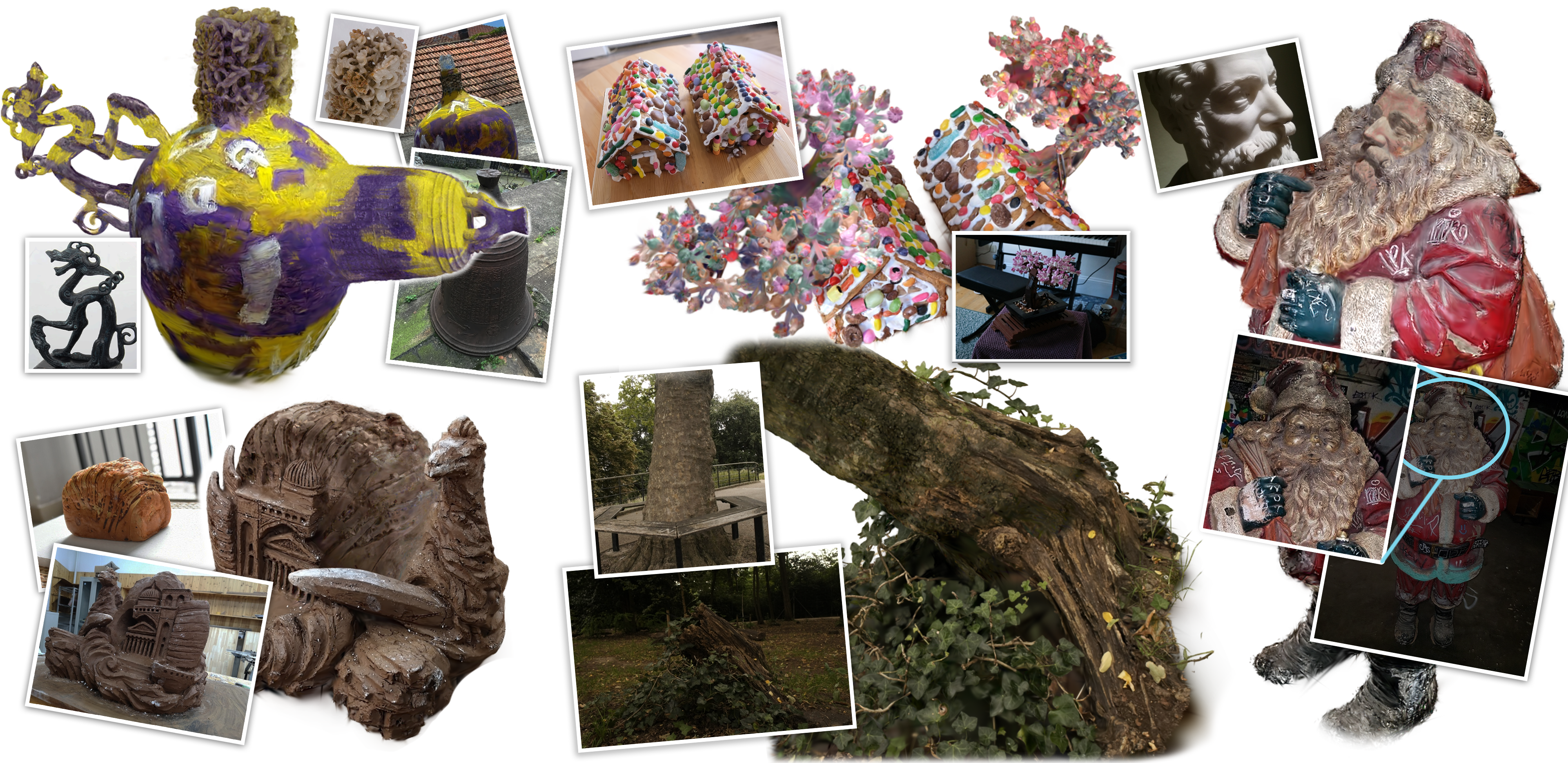}
  \caption{Our method can seamlessly stitch multiple 3D Gaussian fields together \cite{kerbl3Dgaussians} interactively, resulting in new, highly detailed, and realistic objects.
 % that are structurally aware. 
  All of the geometric parts or models are derived from the BlendedMVS \cite{blendedMVS} and Mip360 \cite{barron2022mip360} datasets.}
  \label{fig:teaser}
\end{teaserfigure}


\maketitle

\section{Introduction}

\par
%Neural rendering is a fast-growing field that uses neural techniques to reconstruct \cite{wang2021neus} and render \cite{barron2023zipnerf} high-fidelity 3D models. 
%As we all know, 3D scenes usually consist of many 3D objects, and objects are also composed of many different parts. Compositing different parts from different objects to form new objects is a common way of modeling 3D objects, which is termed example-based modeling\cite{funkhouser2004modeling}. This tool is often used in CG modeling process, where all objects are in CG fashion that are not realistic.
As we all know, 3D scenes typically contain multiple 3D objects composed of various parts. Example-based modeling ~\cite{funkhouser2004modeling} is a technique that involves combining different parts from different objects to create new ones. This is a common tool in Computer Graphics (CG) modeling, where objects are designed in a non-realistic CG fashion.
In this paper, we consider realistic example-based modeling, where all parts are captured from the real world, as shown in Fig. ~\ref{fig:teaser}. This task becomes prominent with the emergence of Neural Radiance Fields, which enables photorealistic 3D reconstruction and rendering.

\par
\begin{figure*}
    \centering
    \includegraphics[width=\textwidth]{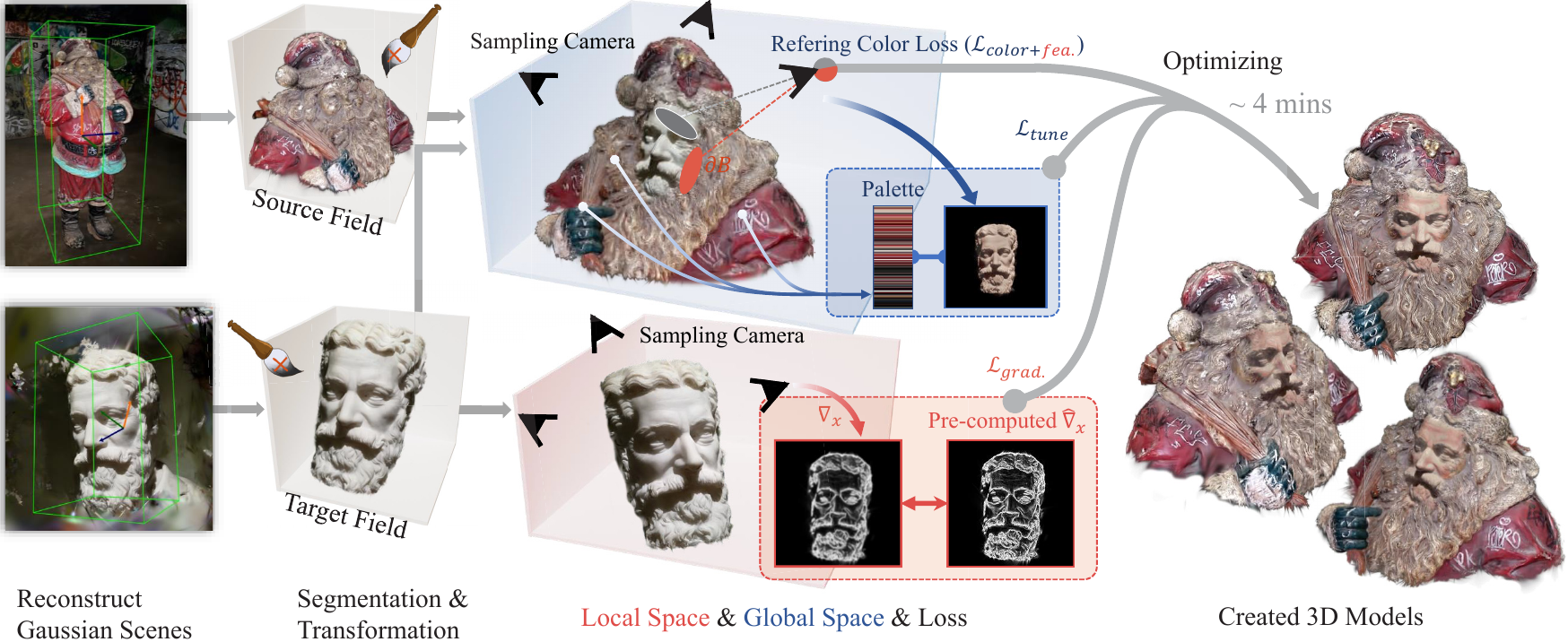}
    \caption{Overview of our framework. Our novel pipeline provides an interactive editing experience and has real-time previewing capabilities to visualize the optimizing process, allowing for the seamless and interactive combination of multiple Gaussian fields. }
    \label{fig:overview}
\end{figure*}

Among the various approaches designed for 3D modeling from multiple neural fields, a portion of the research \cite{Relight3DG2023, liu2023nero} is devoted to the inverse rendering process to achieve consistent lighting and shadowing. But these methods rarely consider a situation where the harmonious and seamless effect is required for merging or unifying two or more neural fields. 
SeamlessNeRF \cite{gong2023seamlessnerf} is the first work to tackle seamless merging, attempting to address the consistency problem by propagating gradients on synthesis cases. Nonetheless, due to its implicit grid-based representation, SeamlessNeRF can neither achieve fine-grained editing (e.g. the face in the \textit{Santa} case in Fig. \ref{fig:overview}) under real-world cases nor provide an interactive workflow in real-time. Additionally, its gradient-based strategy can produce significant artifacts (see Fig. \ref{fig:compare_seamless}) and fails to propagate structural characteristics when the condition becomes more complex (e.g., the \textit{bottle} in the left-upper corner in Fig. \ref{fig:teaser}). Therefore, achieving a harmonious and photorealistic stitching result on real-world data remains an unsolved challenge that needs further exploration.

\par
To address the limitations mentioned above, we propose a new method for interactive editing and stitching multiple parts using explicit shape representation in 3D Gaussian Splatting. Our method has two significant advantages. First, its point-based representation enables fine-grained editing, allowing for detailed appearance optimization and the removal of artifacts. Second, its rasterizer pipeline provides a real-time interactive editing environment.
Due to the discrete and irregular nature of 3D-GS, it is not feasible to conduct gradient propagation as SeamlessNeRF. Thus, 
% a novel sampling-based optimization strategy is proposed, which we found can not only seamlessly propagate color tone but also the structural characteristics.
we introduce a novel sampling-based optimization strategy that can seamlessly propagate not only color tones but also structural characteristics.
Our evaluation benchmarks are primarily derived from real-world scenes, demonstrating our superior ability to handle complex cases.

\par
More specifically, our pipeline takes multiple scenes as input, containing source and target objects represented by 3DGS. We then carefully segment these objects and apply rigid transformations in order to create a semantically meaningful composite in 3D space. An intersection boundary region between the objects is also identified before blending. 
The next is the key step in our process which aims to optimize the appearance of the target objects so that their texture and color match those of the source object.
We achieve this by using a two-phase optimization scheme: the first phase involves sampling-based cloning (S-phase), and the second phase involves clustering-based tuning (T-phase).
During the S-phase, the target field is optimized using a heuristic sampling strategy that considers the structural characteristics at the boundary. Additionally, an efficient 2D gradient constraint is applied to preserve the original texture content of the target field. 
However, optimizing solely with S-phase may lead to the appearance of artifacts or unintended color features that do not fit with the overall composite. 
Therefore, we address this issue with T-phase, where we utilize a pre-calculated feature palette derived from the source field through aggregation and clustering. Subsequently, this palette is applied to tune the target field.
It is important to note that the two-phase optimization is a joint procedure, where losses from the S-phase are always maintained while losses from the T-phase are added later during optimization. 

In summary, our method makes the following contributions:
\begin{itemize}[noitemsep,nolistsep,leftmargin=*]
    \item The first work to use 3D-GS for realistic and seamless part compositing, enabling real-world example-based modeling.  
    %incorporate a realistic and seamless merging effect into 3D-GS.
    % The first work to introduce a realistic and seamless appearance merging effect to 3D GS.
    \item A novel sampling-based optimization strategy is proposed, with which not only the texture color but also the structural characteristics can be propagated seamlessly. 
    % An efficient and effective synthesis method for creating photorealistic 3D models from real-world examples. Using a heuristic sampling strategy, structural characteristics at the boundary can be synthesized and propagated seamlessly.
    % With a heuristic sampling strategy, structural patterns at the boundary can be perceived and propagated seamlessly.
    \item A user-friendly GUI is carefully designed to support an interactive workflow of the modeling process in real time.
    % \item Exhaustive and vivid experiment results show that our method is capable of handling real-world cases across a diverse range of scenarios.
\end{itemize}
\section{Related Work}
%\subsection{Example-based Synthesis}
\subsection{Example-based Seamless Editing}
%Example-based synthesis is a classic modeling and image/texture synthesis technique. 
Seamless editing, particularly in the context of example-based image and texture synthesis, is a well-studied editing technique in computer graphics and image processing.
As for textures, example-based texture synthesis \cite{wei2009state,efros1999texture} intends to seamlessly create textures at any size from exemplars, which has been widely employed in contemporary graphics pipelines and game engines. In 2D image synthesis, patch-based synthesis techniques have been widely researched to seamlessly combine visually inconsistent images \cite{perez2023poisson, darabi2012imagemeld}.
Meanwhile, \citet{Kwatra2005TextureOF} introduced ``Texture Optimization,'' which transfers photographic textures to a target image for example-based synthesis. To facilitate structural image editing tasks, ``PatchMatch'' \cite{Barnes2009PatchMatchAR} found approximate nearest-neighbor correspondences between patches in images for seamless image region reshuffling.
%In terms of 3D models,
In terms of seamless editing in 3D objects, \citet{rocchini1999multiple} and \citet{dessein2014seamless} propose methods for stitching and blending textures on 3D objects, respectively, while \citet{yu2004mesh} use the Poisson equation to implicitly modify the original mesh geometry via gradient field manipulation.
Additionally, example-based modeling can also generate novel models from parts of existing models \cite{funkhouser2004modeling}, allowing untrained users to create interesting and detailed 3D designs, such as city building \cite{merrell2007example}, things arrangements \cite{fisher2012example}, mesh segmentation \cite{katz2005mesh}, and merging \cite{kreavoy2007model}. Recently, deep learning methods have leveraged generative models to generate diverse instances from a single exemplar \cite{wu2022togexample, li2023patchexample} or a cluster of examples \cite{Zhang_2023_CVPR}. Definitely, the example-based methodology is a valuable tool for creating diverse and novel content, which can reduce the workload for the artists or can be leveraged by procedural content generation programs. In our work, we combine this valuable idea with the advanced technique of 3DGS to create content directly from the real world.

\subsection{Neural Scene Composition}
Neural scene composition primarily involves the synthesis of multiple neural objects represented by neural fields, such as free-viewport video \cite{lin2022enerf, zhang2021editablefvv, Wang_2023_CVPR}, autonomous driving \cite{ost2021neural, kundu2022panoptic, tancik2022block, fu2022panoptic, zhou2023drivinggaussian, yang2023unisim} and scene understanding \cite{kerr2023lerf, shuai2022multinb, yang2021learning, wu2022object}.
% For multi-target reconstruction tasks such as free-viewport video \cite{lin2022enerf, zhang2021editablefvv, Wang_2023_CVPR} and driving scene \cite{ost2021neural, zhou2023drivinggaussian, yang2023unisim}, the core idea of those researches is to model a scene as multiple semantic layers \cite{yang2021learning, wu2022object} rather than a whole by cues such as 3D motion \cite{shuai2022multinb} or precomputed segmentation mask \cite{kundu2022panoptic, tancik2022block, fu2022panoptic}.
And for those composition tasks with multiple pre-trained models, mesh scaffold \cite{yang2022neumesh, baskedsdf} or texture extraction \cite{Tang_2023_ICCV, chen2023mobilenerf} from the neural field are preferred to achieve higher render speed or rather fine-grained control. This type of work acts as a ``bridge'' between neural and traditional representations in order to improve performance using the classical graphics pipeline.  
%This kind of works actually serves as a ``bridge''  that combines neural representation with a traditional one to gain better performance with the help of classical graphics pipeline. 
A small portion of the work focuses on creating a mixed render pipeline for neural 3D scene composition tasks, combining traditional render techniques like ray tracing  \cite{qiao2023dynamic}, shadow mapping \cite{gao2023aaaiAGIF}, and ambient occlusion \cite{Relight3DG2023}. There are also a few works that focus on creating a compositional scene with generative models like diffusion models \cite{po2023compositionaldiff}. 
\par
None of those works except Neural Imposter \cite{liu2023neuralimposter} and SeamlessNeRF \cite{gong2023seamlessnerf} focus on example-based modeling by stitching multiple part NeRFs. However, part objects in Neural Imposter are just placed together without any appearance blending, which cannot support a general case of 3D modeling. SeamlessNeRF achieved harmonious results on a small-scale synthesis dataset, making it the first work to discuss seamless example-based modeling with neural techniques today. However, SeamlessNeRF cannot handle real-world cases when the condition becomes more complex, nor can it perform interactive editing, which is commonly required in example-based modeling. On the contrary, our approach overcomes these limitations, performs well in real-world scenarios, and supports interactive editing using Gaussian fields. 

\subsection{3D Gaussians}
3D Gaussian Splatting \cite{kerbl3Dgaussians} is a point-based rendering method that has recently gained popularity \cite{yang2024deformable, huang2024sc, EnVision2023luciddreamer, tang2023dreamgaussian, chen2023gaussianeditor, yang2023gs4d, li2024spacetime} due to its realistic rendering and significantly faster training time than NeRFs. Compared to the implicit representation of NeRF, 3DGS is more advantageous for editing tasks. The superior advance lies in the fact that, unlike previous work that embedded an object in a certain neural field (e.g., learnable grid or MLP network), once clusters of Gaussians are optimized, they can be easily fused together and fed into the rasterizer. The 3DGS pipeline was born with an intrinsic property suitable for composition.
% Its GPU-customized rasterization pipelines and spherical harmonics-based shading enable not only real-time speed but also high quality comparable to NeRF-based methods in both bounded and unbounded scenes. Several concurrent works have adapted 3D Gaussians for dynamic scenes \cite{luiten2023dynamic, yang2023deformable3dgs, yang2023gs4d, huang2023sc-gs, li2023spacetimegaussians}, text-to-3d generation \cite{EnVision2023luciddreamer, tang2023dreamgaussian, chen2023text} and avatars \cite{Zielonka2023Drivable3D, zheng2023gpsgaussian, jiang2023hifi4g}. Furthermore, compared to the implicit representation of NeRF, 3DGS is more advantageous for editing tasks \cite{chen2023gaussianeditor}. The superior advance lies in the fact that, unlike previous work that embedded an object in a certain neural field (e.g., learnable grid or MLP network), once clusters of Gaussians are optimized, they can be easily fused and fed into the rasterizer together. The 3DGS pipeline was born with an intrinsic property suitable for composition.
\section{Seamless Gaussians}
Our approach starts with segmenting interesting parts from pre-trained Gaussian scenes. After acquiring target and source models represented by Gaussians, we carefully transform them to obtain a semantically meaningful composite. Then we optimize the target objects to achieve a harmonious composite through a two-phase (sampling-based cloning and clustering-based tuning) scheme. All these processes can be run interactively and previewed in real-time with our well-tailored GUI design.
\subsection{Segmenting and Transforming Gaussians}
Segmentation is the first step in example-based modeling, which involves picking out interesting parts as the components of the final artwork. Previous works have performed this task by providing guidance using 2D mask \cite{cen2023sa3d, mirzaei2023spin} or injecting semantic label \cite{kerr2023lerf} into a neural field. Now, benefiting from Gaussian representation (resembling point cloud), segmentation can become more practical at a finer-grained level. In our pipeline, we show that a combination of a simple bounding box and a user brush can work very well for a clean mask (see Fig. \ref{fig:gui_segment}). For instance, we can mask the \textit{sculpture} with a brush to match the shape of \textit{Santa}'s face (see Fig. \ref{fig:overview}).

Transformation aims at placing multiple interesting parts $\mathcal{G}_i$ represented by Gaussians to form a semantically meaningful composite $\mathcal{M}$, which can be denoted as:
\begin{equation}
    \mathcal{G}_{i}^{\mathrm{global}} = \normalsize{F}(\mathcal{G}_i^{\mathrm{local}}|\hat{\mathbf{q}}_i,\mathbf{t}_i,s_i)
    , \quad \mathcal{G}_i \in \mathcal{M}
\end{equation}
where $F$ is the rigid transformation applied on one part of Gaussians with rotation $\hat{\mathbf{q}}_i$ (represented in quaternion), translation $\mathbf{t}_i$, and scale $\mathbf{s}_i$, transforming the part from its local space to the global space. Specifically, the partial attributes of each $\mathcal{G}$ should be modified, which includes position $\mathbf{x}$, scaling $\mathbf{s}$, rotation $\mathbf{q}$ (in quaternion), and feature $\mathbf{f}$ (represented as spherical harmonics). The position and scaling can be performed trivially, while the transformed rotation $\mathbf{q}'$ and feature $\mathbf{f}'$ can be expressed as:
\begin{equation}
    \begin{aligned}
        \mathbf{q}' &=  \mathbf{q}\hat{\mathbf{q}}, \\
        \mathbf{f}' &= M_{bands}(\mathbf{f} \: | \: \hat{\mathbf{q}}),
    \end{aligned}
\end{equation}
where $M_{bands}$ means we use a set of matrices to rotate each band of SH coefficients introduced by \cite{ivanic1996shrotation}.

\begin{figure}
    \centering
    \includegraphics[width=\linewidth]{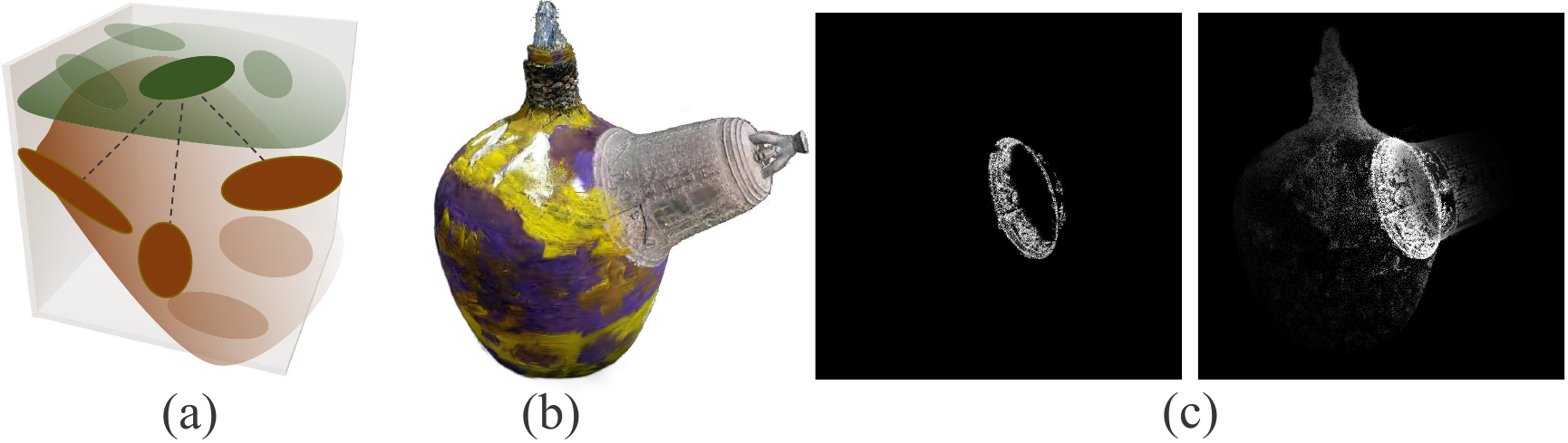}
    \caption{For a Gaussian point in the target field, its (a) K-nearest neighbors in the source field can be leveraged to justify whether this point belongs to the intersection boundary region. We use the boundary of (b) as an example to demonstrate the effectiveness of this strategy, as shown in (c).}
    \label{fig:border-identification}
\end{figure}

\subsection{Boundary Condition by KNN Analyzing}
After transformation, certain points in one field approach another field (see Fig. \ref{fig:border-identification}), forming intersection boundary regions between all Gaussians. For the sake of simplicity, we will use two Gaussians, source field and target field, to demonstrate our approach.
% After transformation, certain points in one field get close to another field (Fig. \ref{fig:border-identification}), forming boundary regions between all Gaussians. For simplicity, we demonstrate our approach with two Gaussians, named source field and target field. 

%Before optimization, the boundary points should be identified in the target field, which is the crucial and initial condition for harmonizing. 
Before optimization, the boundary points in the target field must be identified, as this is the critical and initial condition for harmonization.
For each Gaussian point in target field $\mathcal{T}$, we search its K-nearest neighbors in source field $\mathcal{S}$, which can be denoted by:
\begin{equation}
    \left\{ b_i \right\}_K = \mathop{KNN}\limits_{\mathcal{S}}(a), \quad a \in \mathcal{T},b_i \in \mathcal{S}
\end{equation}
where $a$ is a point in the target field, and $b_i$ is a point in the source field. Whether a point $a$ belongs to boundary $\partial B$ can be identified as $a \in \partial B$ iff.:
\begin{equation}
    \frac{1}{K}\sum\limits_{i}^{K}|b_i-a| < \beta \quad \mathrm{and} \quad o(a)>\tau,
\end{equation}
where $o(a)$ is the opacity of that Gaussian point, $|b_i-a|$ is the Euclidean distance between $b_i$ and $a$. $\tau$ and $\beta$ are thresholds and we empirically set $\tau$ to 0.95, $\beta$ to $0.05 \times L$. $L$ is the size of the composite. (e.g. measured by the bounding box).
An additional method for a better boundary condition on real-world data is that we discard outliers in both fields (e.g. some Gaussian points are far from the others, which may occur in some scenes).

% As for these boundary points, we calculate referenced features for them such as to confirm the boundary condition. 
We calculate referenced features for these boundary points in order to confirm the boundary condition. For each $a \in \partial B$, its target feature is:
\begin{equation}
    \hat{\mathbf{f}}(a) = \frac{1}{K}\sum\limits_{i}^{K}\mathbf{f}'(b_i), \quad a \in \partial B, b_i \in \mathop{KNN}\limits_{\mathcal{S}}(a)
\end{equation}
where $\mathbf{f}'(b_i)$ is the feature of $b_i$ after transformation. 
%To reach this boundary condition is to optimize boundary points towards their target features:
To achieve this boundary condition, we optimize boundary points toward their target features:
\begin{equation}
    \mathcal{L}_{feature} = \sum\limits_{a \in \partial B} \left\lVert \mathbf{f}'(a) - \hat{\mathbf{f}}(a) \right\rVert_{2}^{2},
\end{equation}
where $\mathbf{f}'(a)$ is the feature of $a$ and we directly apply this loss on SH coefficients.

\begin{figure}[tbp]
    \centering
    \begin{minipage}{\linewidth}
        \includegraphics[width=\linewidth]{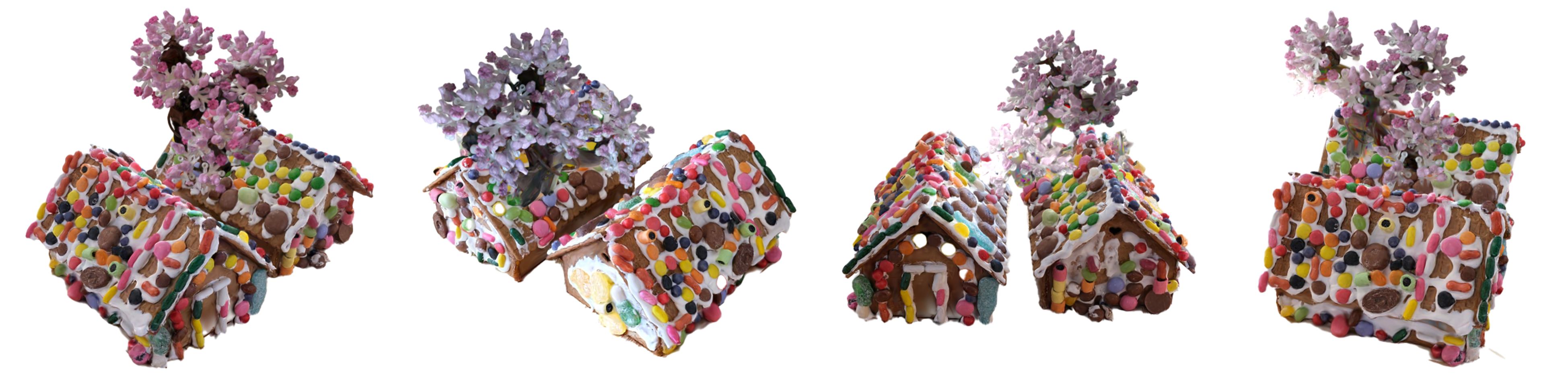}
        \includegraphics[width=\linewidth]{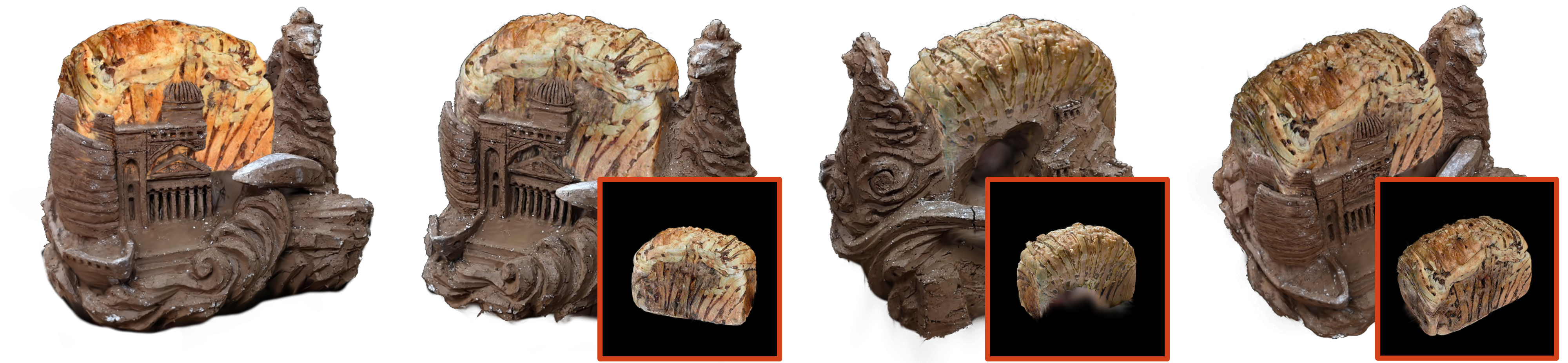}
    \end{minipage}
    \caption{ablation study on the color loss in the S-phase. 
    Without color loss, the propagation is inefficient and will not begin. The cases shown above have been running for more than twice as long, but they are still trapped in insufficient propagation. It is because, without color loss, only a small number of points' features need to be updated at first, as opposed to shared weights in an MLP applied to all points. That minor ``forces'' cannot drive the overall minimization of the gradient loss.}
    % Without color loss, the propagation is inefficient and even won't start. The cases above have been running over 2 times longer but are still trapped in insufficient propagation.
    \label{fig:wont_start}
\end{figure}

\begin{figure*}[htbp]
    \centering
    \includegraphics[width=\linewidth]{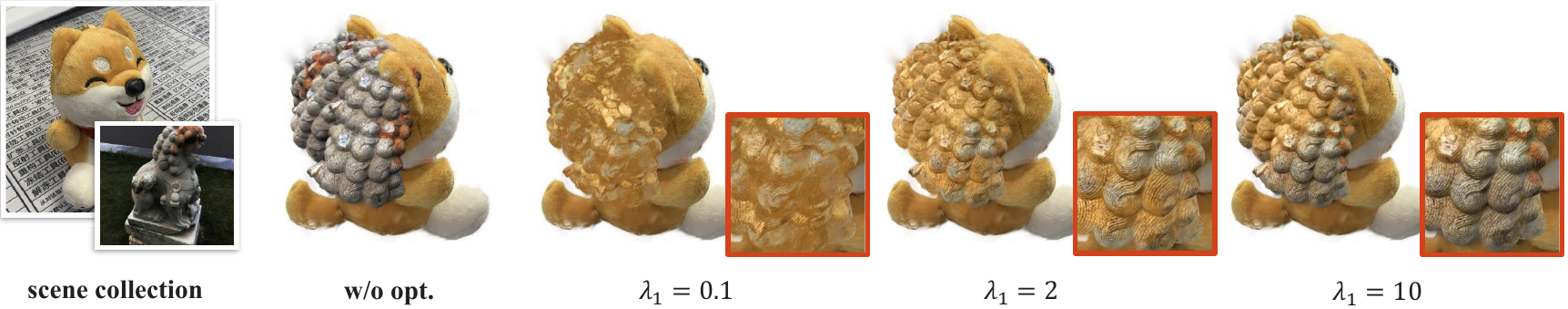}
    \vspace{-0.6cm}
    \caption{ablation study on the effectiveness of gradient loss for different weights. Experiments show that higher weights can help to preserve more content while preventing harmonization. }
    \label{fig:ablation_gradweight}
\end{figure*}
\begin{figure}[htbp]
    \centering
    \includegraphics[width=\linewidth]{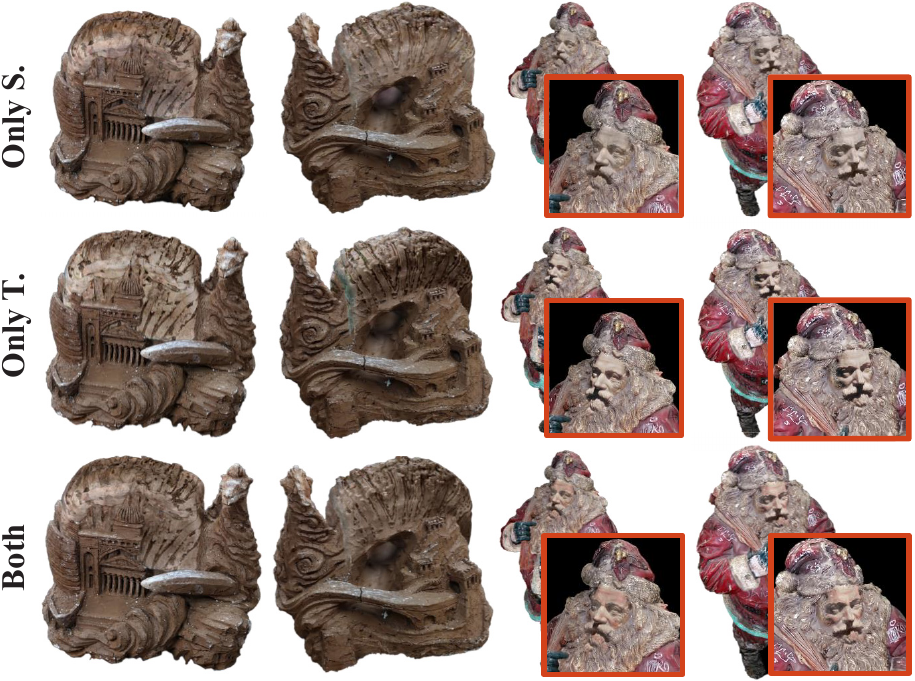}
    \caption{ablation study on sampling-based cloning (S.) and clustering-based tuning (T.). Here, ``Both'' means the full scheme.}
    \label{fig:ablation_local_global}
\end{figure}
\begin{figure}[tbp]
    \centering
    \includegraphics[width=\linewidth]{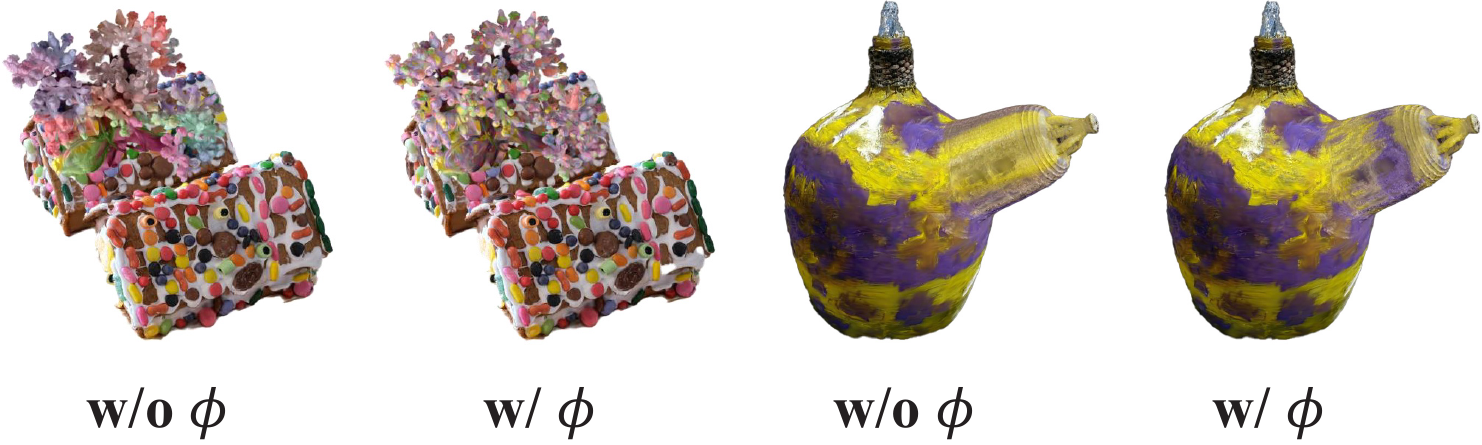}
    \caption{Ablation study on the impact of mapping function $\phi$ in the S-phase.
    % ablation study on impact of mapping function $\phi$ in S-phase. 
    Random effects make composition more realistic.}
    \label{fig:sample_graph_warp}
\end{figure}

\subsection{Sampling-based Cloning}
We propose sampling-based cloning as our ``S-phase'' in optimization. The core idea is how to seamlessly propagate the style in boundary through the remaining points in the target field while preserving its rich content. In contrast to a regular grid suitable with a gradient-based strategy in SeamlessNeRF \cite{gong2023seamlessnerf}, Gaussian points are irregularly and discretely distributed in 3D space. As a result, alternative approaches need to be explored. A straightforward idea is that given a point in target field $\mathcal{T}$, one can calculate the feature difference between that point and its neighbors in $\mathcal{T}$, resembling ``Laplacian coordinates''. Then, one can use that ``difference'' as the regularizer while minimizing $\mathcal{L}_{feature}$. 
% However, this naive approach may encounter failure (see Fig. \ref{fig:wont_start}) even before the commencement of propagation, primarily because only a limited number of points' features need updating at the beginning, contrasting the shared weights in an MLP applied to all points. 
However, this naive approach may fail even before propagation begins  (see Fig. \ref{fig:wont_start}).
% owing to the fact that only a small number of points' features need to be updated at first, as opposed to shared weights in an MLP applied to all points.
% As a result, such minor ``forces'' cannot drive the overall minimization of the regularizer. 
Furthermore, the boundary's structural characteristics (such as the \textit{bottle}-\textit{bell} intersection in the right-upper corner of Fig. \ref{fig:compare_seamless}) necessitate seamless cloning, which significantly improves the stitching quality. 
%Consequently, such minute ``forces'' cannot drive the comprehensive minimization of the entire regularizer. Moreover, the structural characteristics (e.g. the \textit{bottle}-\textit{bell} intersection in the right-upper corner in Fig. \ref{fig:compare_seamless}) at the boundary require seamless cloning, a process that notably enhances quality.
\par
Hence, we propose an effective sampling strategy to explicitly propagate features for each remaining point outside the boundary. The core idea lies in the way of searching several ``driven points'' for a candidate. The color of the candidate is driven by those points. For each $a \in \mathcal{T}-\partial B$, the optimizing target of its color in direction $\mathbf{d}_a$ is:
\begin{equation}
    \hat{\mathbf{f}}(a,\mathbf{d}_a) = \frac{1}{K}\sum\limits_{i}^{K}\mathbf{f}'(b_i,\mathbf{d}_b), \quad a \in \mathcal{T}-\partial B, \: b_i \in \mathop{KNN}\limits_{\partial B}(\phi(a))
    \label{eq:sampling_strategy}
\end{equation}
where $\mathbf{f}(a,\mathbf{d}_a)$ means sampling SH color in view direction $\mathbf{d}_a$ (from point $a$ to camera), the same as $\mathbf{f}(b,\mathbf{d}_b)$. 
The camera centers are uniformly sampled from the surface of a sphere centered on the composite object's origin.
%It's worth noting that $KNN(\phi(a))$ constitutes the strategy responsible for the structural characteristic, which is designed to assign all ``driven points'' correlated continuously in space for all corresponding candidate points within a small neighborhood. 
It is important to note that the sampling strategy $KNN(\phi(a))$ maps the locations of nearby candidate points $a$-s  to the correlated neighboring ``driven points'' and inherits the continuity of the textures from those ``driven points''.
We use $\phi(x)= x + sin(\gamma \cdot \delta x)$ to add random effect by disturbing KNN searching (see Fig. \ref{fig:sample_graph_warp}), where $x$ is the position of $a$, $\delta x$ is the distance between $a$ and its nearest $b_i$ in boundary, and $\gamma$ is empirically set to 10. A larger $\gamma$  is suitable for higher structural frequencies.
% It is worth noting that $\phi$ is a mapping function that contributes to the structural characteristic, and we use $\phi(x)= x + sin(\gamma \cdot \delta x)$ to add this effect by disturbing KNN searching (see Fig. \ref{fig:sample_graph_warp}), where $x$ is the position of $a$, $\delta x$ is the distance between $a$ and its nearest $b_i$ in boundary, and $\gamma$ is empirically set to 10. A larger $\gamma$  is suitable for higher structural frequencies. 
In this way, we can synthesize structurally aware stitching results. With Eq.~\eqref{eq:sampling_strategy}, we add a color loss to the S-phase:
\begin{equation}
    \mathcal{L}_{color} = \sum\limits_{a \in \mathcal{T}-\partial B} \left\lVert \mathbf{f}'(a,\mathbf{d}_a) - \hat{\mathbf{f}}(a,\mathbf{d}_a) \right\rVert_{2}^{2},
\end{equation}
so that the color of those candidates can be optimized towards their target to achieve our explicit feature propagation.

\par
To preserve the original rich content in $\mathcal{T}$, we present a more efficient gradient loss calculated in the local space of $\mathcal{T}$, leveraging the guidance in 2D screen space:
\begin{equation}
\begin{aligned}
    \mathcal{L}_{grad} &= \sum\limits_{x \in I} \left\lVert \nabla_{x}I^{\mathcal{T}}(p) - \hat{\nabla}_{x}I^{\mathcal{T}}(p) \right\rVert_{2}^{2}, \\
    I^{\mathcal{T}}(p) &= \mathcal{R}(\mathcal{G}_{\mathcal{T}}^{local},p),
\end{aligned}
\end{equation}
where $p$ is the randomly sampled camera in the local space of target field $\mathcal{T}$, $I$ is the rendered color image by rasterizer $\mathcal{R}$ of 3DGS. We pre-calculate $\hat{\nabla}_{x}I$ for each camera with the Sobel operator \cite{sobel19683x3} before the optimization starts. We found that supervising gradients in screen space is more efficient than the straightforward one, as shown in Fig. \ref{fig:ablation_knnsobel}.

\begin{figure*}
    \centering
    \includegraphics[width=\linewidth]{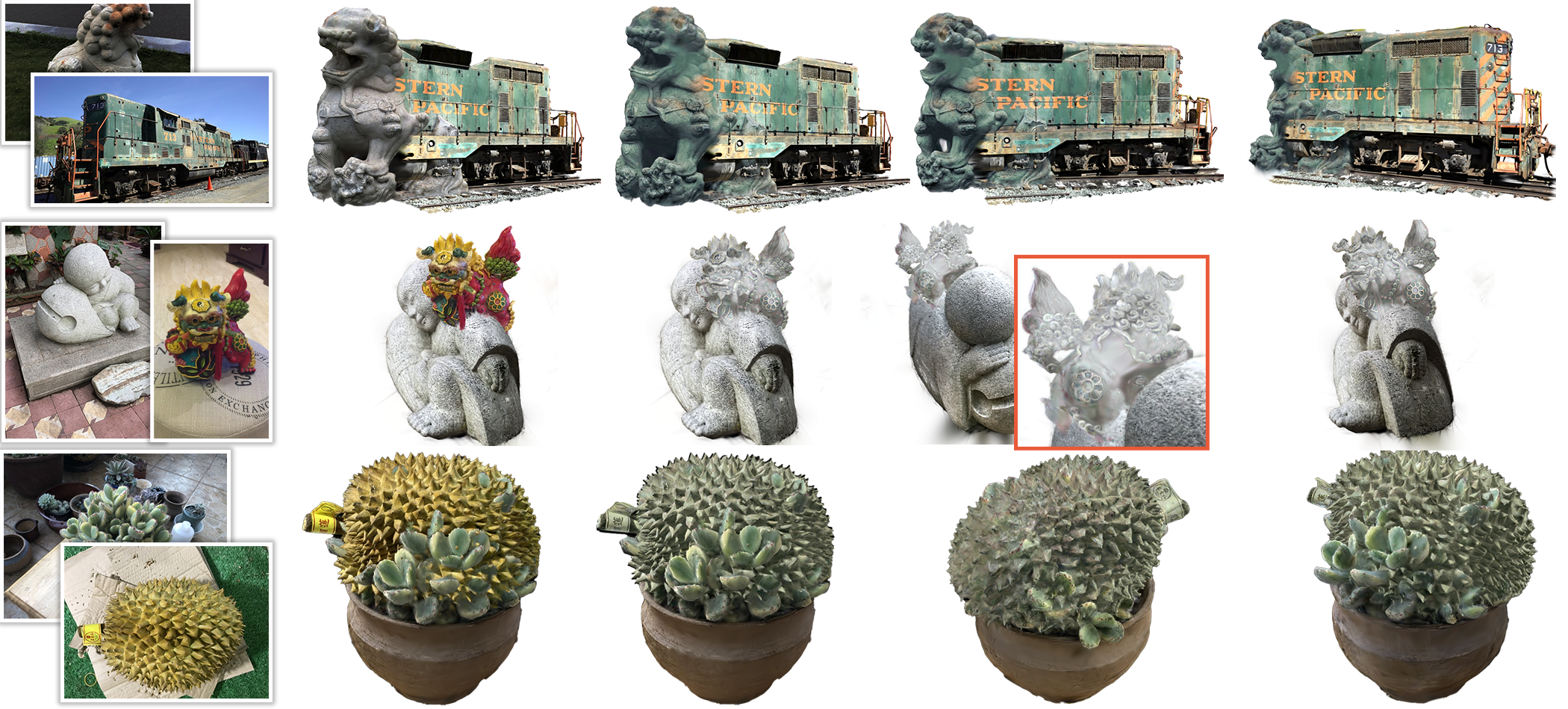}
    \vspace{-0.6cm}
    \caption{Results for more real-world data from the BlendedMVS \cite{blendedMVS} and Mip360 \cite{barron2022mip360} datasets, demonstrating that our method can produce realistic effects in real-world scenarios.}
    \label{fig:main_res}
\end{figure*}

\subsection{Clustering-based Tuning}
% Optimizing only with S-phase can cause artifacts or unintended color features w.r.t. the whole composite. 
%Optimizing only with S-phase can result in artifacts or unintended color features across the entire composite.
%Thus, we propose clustering-based tuning as our ``T-phase'' in optimization, which enhances the overall harmony of the composite. The T-phase performs similarly to a palette. We first aggregate and cluster (e.g., k-means) the color of the source field from various angles:
While S-phase optimization is effective in preserving local color consistency, relying solely on it may lead to misaligned global appearance, such as uneven brightness, hues, and saturation (See Fig. \ref{fig:ablation_local_global}). Therefore, we propose using a clustering extracted color palette to perform global tuning, which we refer to as the "T-phase" in optimization. This approach enhances the overall harmony of the composite by performing dynamic matches to a palette. To implement the T-phase, we first aggregate and cluster the color of the source field from various angles:
% using techniques such as k-means
% Thus we propose clustering-based tuning as our ``T-phase'' in optimization, which further promotes the harmony of the whole composite. The T-phase performs resembling a palette. We first aggregate and cluster (e.g. k-means) the color of the source field from a bunch of directions:
\begin{equation}
    \{\mathbf{c}_i\}_N, \{w_i\}_N \leftarrow \mathcal{A}(\mathcal{G}_{\mathcal{S}}^{global}),
\end{equation}
where $\mathbf{c}_i$ is the color (cluster center) in palette, $w_i$ is the sample percentage occupied by the center, and $\mathcal{A}$ stands for our aggregation algorithm. 
%Inspired by Li et al. \cite{li2023compressing}, we use a streaming approach to accelerate the entire aggregation process, considering the large number of samples from the source field. First, we initialize the palette with three bins, each representing a color center, and then start iterating. During each iteration, we collect a set of color samples from a random view. Each color sample votes for the bin closest to them. A new color center for each bin is formed by averaging the original color center and the new color samples collected in that bin. Specifically, for color samples that are far from the existing centers, the bins will expand to accommodate them by assigning a new center to those samples. A color center will expire if there aren't enough samples to vote for it after 20 iterations. The process is repeated until all color centers are almost stable.
Our approach, inspired by Li et al.'s work \cite{li2023compressing}, uses a streaming method to accelerate color aggregation. We start with three bins, collect color samples from a random view, and calculate the new color center for each bin by averaging the original center and new samples collected in it. The number of bins expands to accommodate far-off samples. Centers expire after 20 iterations with no sufficient votes. We repeat this process until all color centers are stable.
% During each single iteration, we collect a batch of color samples from a random view. Every color samples vote to the bin nearest to them. By averaging the origin color center and new color samples collected in each bin, a new color center for each bin forms. Especially, for those color samples far from the existing centers, the bins will grow to fit them by assigning a new center using those samples. A color center will expire if there are not enough samples to vote for it every 20 iterations. The step repeats until all color centers are almost stable.
\par
Once the aggregation process finishes, those color centers will form a palette (see Fig. \ref{fig:overview}). We employ the following loss in our T-phase as a pixel-wise summation:
% \begin{equation}
% \begin{aligned}
%     \mathcal{L}_{tune} &= \sum\limits_{x \in I'} w_{\chi}\left\lVert I^{\mathcal{T}}_{x}(p) - \mathbf{c}_{\chi} \right\rVert_{2}^{2}, \\
%     I' &\leftarrow \{x|x \in I \: \mathrm{and} \: \alpha(x)>0.95\}, \\
%     \chi &= \mathop{\arg\min}\limits_{1 \leq i \leq N}\left\{\left\Vert \mathbf{c}_{i} - I^{\mathcal{T}}_{x}(p) \right\Vert_{2} - w_{i}\right\},
% \end{aligned}
% \end{equation}
%

\begin{equation}
\begin{aligned}
    \mathcal{L}_{tune} &= \sum\limits_{\mathbf{c} \in I'} w_{\chi_c}\left\lVert \mathbf{c} - \mathbf{c}_{\chi_c} \right\rVert_{2}^{2}, ~
    I' \leftarrow \{ I^\mathcal{T}_x(p) | \alpha(x)>0.95\}, \\
    \chi_c &= \mathop{\arg\min}\limits_{1 \leq i \leq N}\left\{\left\Vert \mathbf{c} - \mathbf{c}_{i} \right\Vert_{2} - w_{i}\right\},
\end{aligned}
\end{equation}
where $p$ is the randomly sampled camera in the global space, and $\alpha$ is the alpha mask corresponding to $I^{\mathcal{T}}$.  Both $\alpha$ and $I^{\mathcal{T}}$ are rendered by rasterizer $\mathcal{R}$. $\chi$ represents the target bin’s index, and it is determined by both the distance from color centers and the probability density of bins. Our final total loss function can then be expressed as:
\begin{equation}
    \mathcal{L}_{total} = \mathcal{L}_{feature} + \mathcal{L}_{color} + \lambda_1\mathcal{L}_{grad} + \lambda_2\mathcal{L}_{tune},
\end{equation}
where both $\lambda_1$ and $\lambda_2$ are empirically set to 2 in our experiments.
\section{experiment}
To test the effectiveness and generality of our approach, we conducted experiments on a variety of fascinating 3D objects. 
% We interactively built 21 compositing results, which made up of totally 39 part models: 17 are from BlendedMVS \cite{blendedMVS}, 4 are from Mip360 \cite{barron2022mip360}, 16 are from seamlessNeRF datasets, and 2 are made by ourselves in graphics engine. 
% For more results or the implementation details, refer to our supplementary due to 7-page length.
We interactively built 21 composite results, comprising a total of 39 part models: 17 from BlendedMVS \cite{blendedMVS}, 4 from Mip360 \cite{barron2022mip360}, 16 from SeamlessNeRF datasets, and 2 created by ourselves in a graphics engine.
For more results or the implementation details, please refer to our supplementary materials.

\subsection{Qualitative Comparison}
\begin{figure}[tbp]
    \centering
    \includegraphics[width=\linewidth]{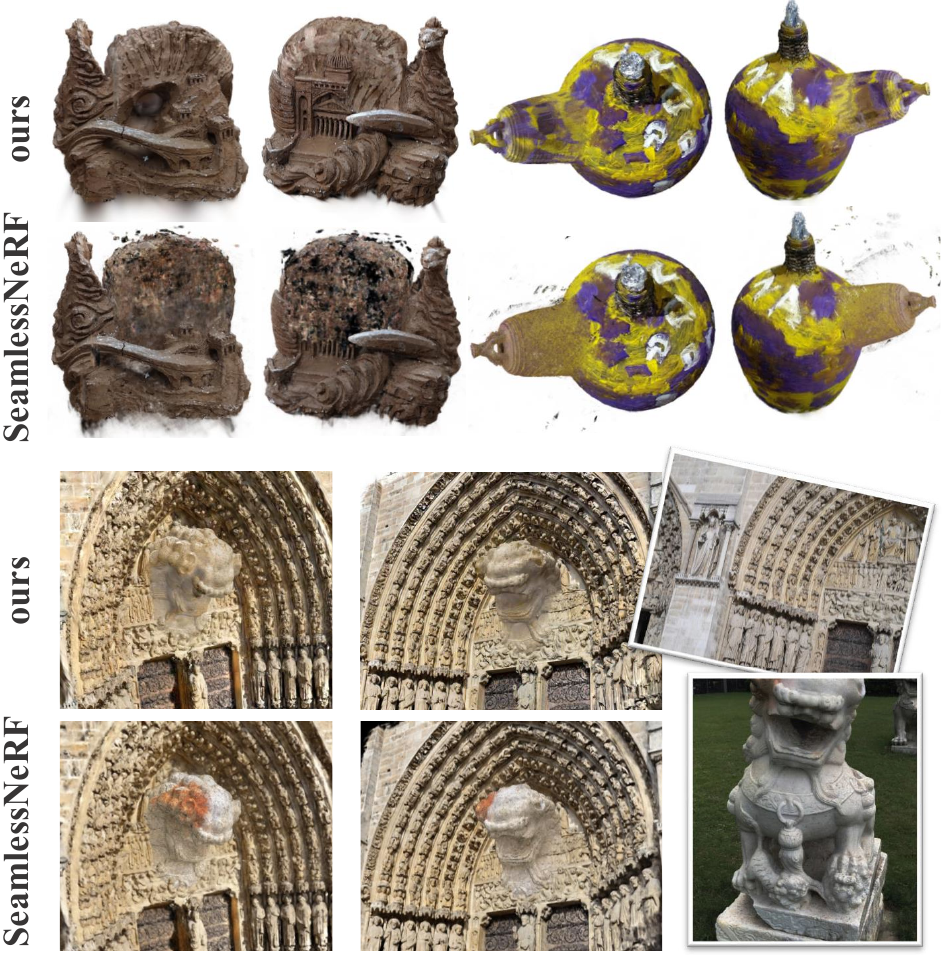}
    \vspace{-0.6cm}
    \caption{Comparisons between our approach and the baseline methods \cite{gong2023seamlessnerf}. SeamlessNeRF failed in all of these real-world scenarios.
    %Comparisons between our approach and baseline methods \cite{gong2023seamlessnerf}. The SeamlessNeRF has failed in these real-world cases.
    }
    \vspace{-0.3cm}
    \label{fig:compare_seamless}
\end{figure}

\begin{figure}[tbp]
    \centering
    \includegraphics[width=\linewidth]{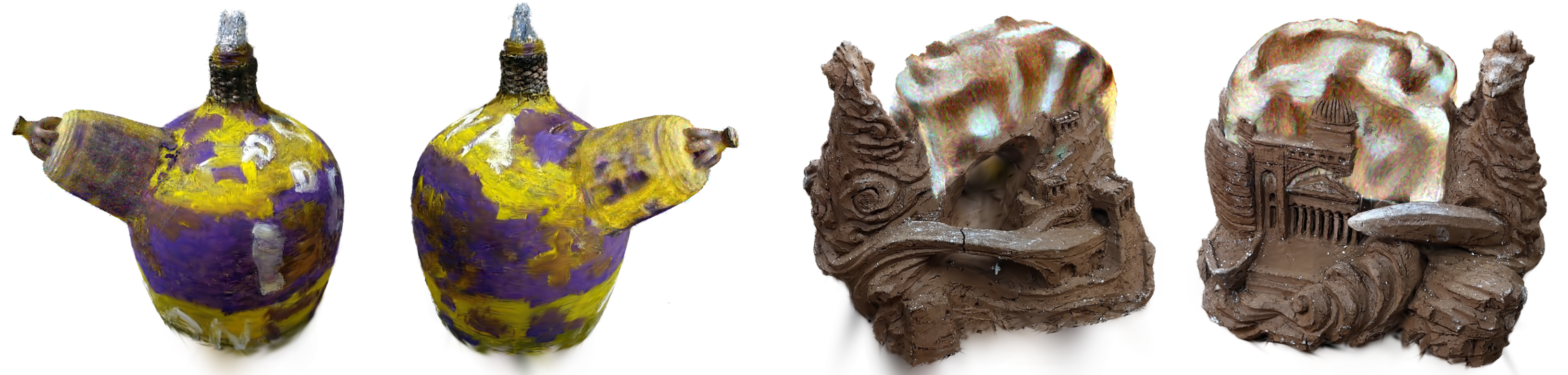}
    \caption{We show that these style-transfer results \cite{meta2022snerf} fail to achieve our effect. Here, we re-implement SNeRF's strategy \cite{meta2022snerf} based on Gaussians to produce results above.}
    \label{fig:compare_snerf}
\end{figure}

We compare our method to SeamlessNeRF \cite{gong2023seamlessnerf}, the first and most recent work that approaches our goal. Fig. \ref{fig:compare_seamless} depicts three comparison cases. In the first case (clay \& bread), SeamlessNeRF failed to achieve high-level geometry editability and struggled with artifacts caused by implicit representation. In the second case (bottle and bell), SeamlessNeRF failed to maintain a harmonious seamless effect due to applying the gradient-based strategy on the complex boundary. In the third case, SeamlessNeRF failed to propagate sufficient color tones due to the complex gradients in the boundary. In addition, we show that the 2D-guided style-transfer method \cite{meta2022snerf} cannot produce a seamless stitching effect, as shown in Fig. \ref{fig:compare_snerf}.  On the contrary, ours can handle all of these situations while producing harmonious results.
\subsection{Quantitative Comparison}
Currently, there is neither a specialized dataset providing ground truth nor an established metric to assess the realism of a 3D model's appearance, making it challenging to evaluate the effectiveness of our approach quantitatively. Nevertheless, we force an evaluation utilizing VQA(Video Quality Assessment) methods, as outlined by \citet{wu2023dover}, and explored the use of 2D projection in video display for assessment purposes. Our results, presented in Tab. \ref{tab:VQA_total}, demonstrate that our average score surpasses the baseline. For a comprehensive understanding of the quantitative experiments, please refer to our supplementary materials.
\begin{table}[tp]
    \centering
    \resizebox{0.8\linewidth}{!}{
    \begin{tabular}{l|cc}
    \hline
              & \multicolumn{1}{l}{ours} & \multicolumn{1}{l}{SeamlessNeRF} \\ \hline
    VQA average score $\uparrow$ & 0.784                    & 0.753                            \\ \hline
    \end{tabular}
    }
    \caption{Quantitative comparison between ours and the baseline.}
    \label{tab:VQA_total}
\end{table}

\subsection{Ablation Study}
\begin{figure}[tp]
    \centering
    \includegraphics[width=\linewidth]{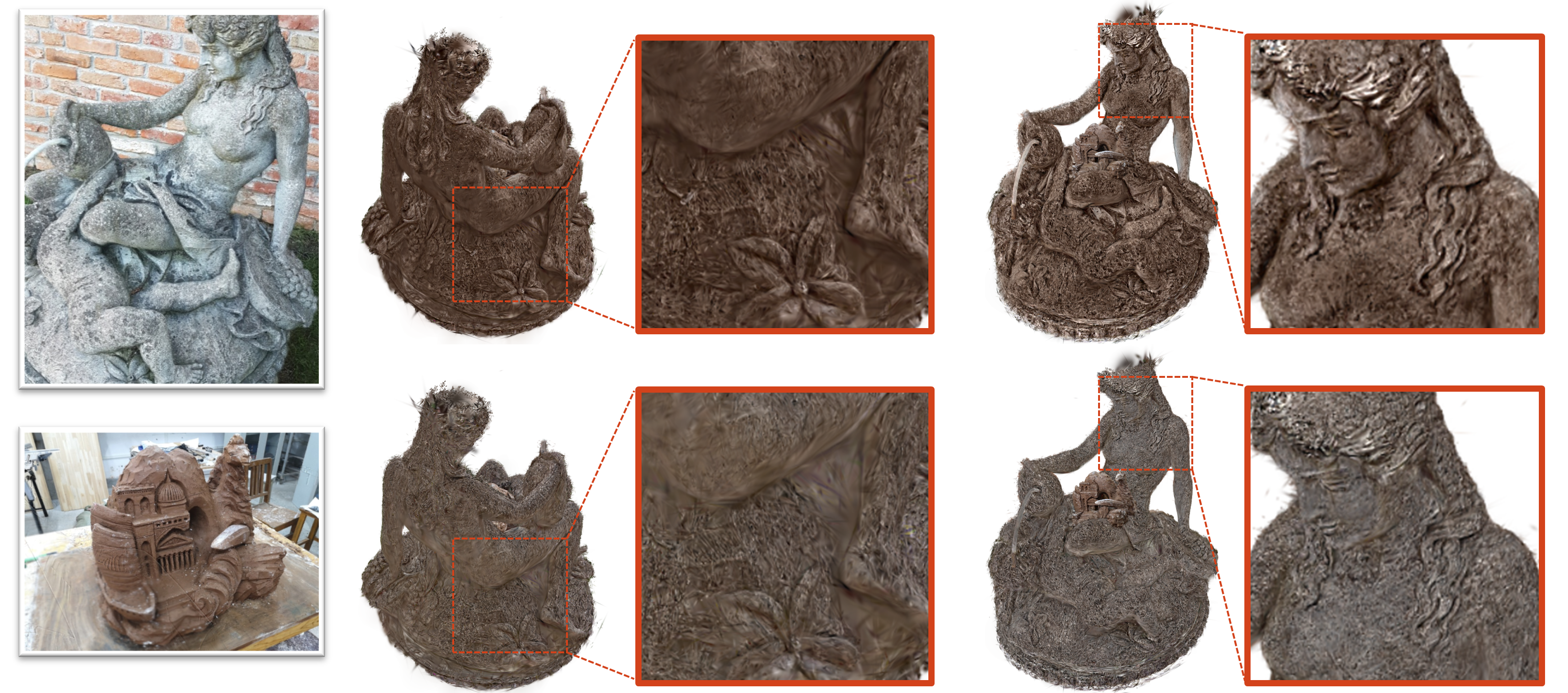}
    \caption{ablation study on two kinds of gradient loss. The 2D gradient supervision (upper row) is more effective than the straightforward one since it focuses on the surface instead of the whole space.}
    \label{fig:ablation_knnsobel}
\end{figure}

\begin{figure}[tbp]
    \centering
    \includegraphics[width=\linewidth]{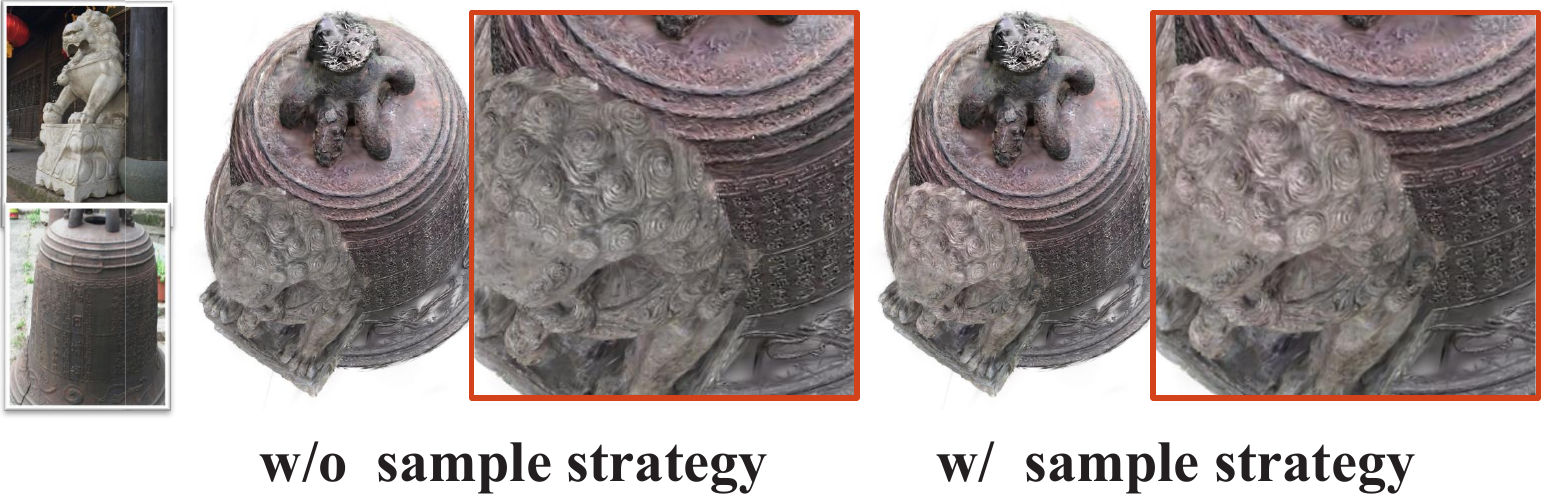}
    \caption{Ablation study on keeping view-dependent effects by sampling strategy in the S-phase. With that strategy used in Equ. \eqref{eq:sampling_strategy}, the upper-view color of paint on \textit{bell} is properly propagated.}
    \label{fig:ablation_sampleSH}
\end{figure}

\paragraph{Effectiveness of 2D Gradient Loss.} 
Fig. \ref{fig:ablation_gradweight} depicts the effect of gradient loss at various weights. Higher weights can help to preserve more content while obstructing harmonization. Fig. \ref{fig:ablation_knnsobel} demonstrates that 2D gradient loss with Sobel operator is significantly more effective than the simple one mentioned in Sec. 3.3.
% Fig. \ref{fig:ablation_gradweight} manifest the effect of gradient loss with varying weights, the higher weight can be helpful with preserving more content but obstructive to harmonization. Fig. \ref{fig:ablation_knnsobel} shows that 2D Gradient Loss with Sobel operator is much more effective than that straightforward one mentioned in sec. 3.3.
% even though the weight of the latter is already set over 50.
\paragraph{Functionality of S-phase and T-phase.} We demonstrate the efficacy of our two-phase scheme in Fig. \ref{fig:ablation_local_global}. 
The S-phase aids in seamless boundary formation, while the T-phase aids in global harmonization when only the S-phase is present.
% The S-phase is helpful with seamless boundary, and the T-phase is helpful with global harmonizing when only based on the S-phase.
\paragraph{Effectiveness of Sampling Strategy for View-dependent Effects.} We ablate the sampling strategy in the S-phase (see Fig. \ref{fig:ablation_sampleSH}) to show that view-dependent effects can be properly propagated using this strategy instead of random sampling.
% We also ablate the sampling strategy in S-phase (see Fig. \ref{fig:ablation_sampleSH}) to demonstrate that the view-dependent effects can be properly propagated with the sampling strategy rather than random sampling.


\subsection{Editor and Application}
To enable a practical and user-friendly workflow, we created an interactive GUI editor that can control and visualize any procedure in the entire process in real-time, including Gaussian segmentation and transformation, boundary identification, and optimization (see Fig. \ref{fig:visualize_gradually_opt} and refer to the supplementary video for more details). Our framework can generate high-fidelity and seamless results across a wide range of real-world scenarios, providing distinct advantages in the direct creation of imaginative 3D models from reality.
% To enable a practical and user-friendly workflow, we built an interactive GUI editor to control and visualize any procedure in the whole process, such as segmenting and transforming Gaussians, identifying boundaries, and optimization (refer to supplementary video for more details). Our framework can efficiently generate high-fidelity and seamless results across a diverse range of real-world scenes, presenting distinct advantages in the direct creation of imaginative 3D models from reality.
\section{Conclusions and Limitations}
We have developed a highly efficient and effective interactive framework for creating realistic 3D models. The method involves stitching Gaussian components seamlessly to create a harmonious 3D model that is an accurate representation of the real world. Our approach has been tested on real-world datasets and has proved to be capable of handling complex cases with a user-friendly interface. This presents a promising avenue for example-based modeling directly from the real world.
% We presented an effective and efficient framework for creating realistic 3D models by harmoniously stitching Gaussian components, demonstrating a promising avenue for example-based modeling directly from the real world. Experiments have shown that our approach can handle complex cases on real-world datasets. Furthermore, with an interactive GUI, users can quickly create 3D models and preview them in real time.

\paragraph{Limitations and Future Work.} 
% Currently, the work can fail in those cases where the target field or source field has intense illumination and largely varying material properties. Another issue is that the absence of certain global features (e.g. the white paint dots in \textit{clay sculpture} case) still remains due to a kind of much finer-grained detail in texture content. Introducing both inverse rendering techniques and style-transfer methods to handle these challenging cases is an interesting topic for future work. 
Currently, our work is unable to transform Gaussian models in a non-rigid manner, which may make it difficult to develop more imaginative cases. To enable a more flexible composition, we can use deformation methods such as \textit{ARAP} \cite{igarashi2005rigid} in the future.  Furthermore, achieving a consistent lighting effect can help improve composition quality under intense lighting.
% Currently, our work doesn't handle transforming Gaussian models in a non-rigid manner, which may be inconvenient to create any more imaginative cases. To enable a more flexible composition, we can employ any deformation rules such as \textit{ARAP (as-rigid-as-possible)} in the future. In terms of further enhancing the composition quality under intense lighting cases, achieving a consistent shading effect can also be an interesting topic for future work.

% \begin{acks}
% \end{acks}

\bibliographystyle{ACM-Reference-Format}
\bibliography{sample-base}


\begin{thebibliography}{60}

%%% the Web address.

\ifx \showCODEN    \undefined \def \showCODEN     #1{\unskip}     \fi
\ifx \showDOI      \undefined \def \showDOI       #1{#1}\fi
\ifx \showISBNx    \undefined \def \showISBNx     #1{\unskip}     \fi
\ifx \showISBNxiii \undefined \def \showISBNxiii  #1{\unskip}     \fi
\ifx \showISSN     \undefined \def \showISSN      #1{\unskip}     \fi
\ifx \showLCCN     \undefined \def \showLCCN      #1{\unskip}     \fi
\ifx \shownote     \undefined \def \shownote      #1{#1}          \fi
\ifx \showarticletitle \undefined \def \showarticletitle #1{#1}   \fi
\ifx \showURL      \undefined \def \showURL       {\relax}        \fi
% invisible to TeX
\providecommand\bibfield[2]{#2}
\providecommand\bibinfo[2]{#2}
\providecommand\natexlab[1]{#1}
\providecommand\showeprint[2][]{arXiv:#2}

\bibitem[Barnes et~al\mbox{.}(2009)]%
        {Barnes2009PatchMatchAR}
\bibfield{author}{\bibinfo{person}{Connelly Barnes}, \bibinfo{person}{Eli Shechtman}, \bibinfo{person}{Adam Finkelstein}, {and} \bibinfo{person}{Dan~B Goldman}.} \bibinfo{year}{2009}\natexlab{}.
\newblock \showarticletitle{PatchMatch: A Randomized Correspondence Algorithm for Structural Image Editing}.
\newblock \bibinfo{journal}{\emph{ACM Trans. Graph.}} \bibinfo{volume}{28}, \bibinfo{number}{3}, Article \bibinfo{articleno}{24} (\bibinfo{date}{jul} \bibinfo{year}{2009}), \bibinfo{numpages}{11}~pages.
\newblock
\showISSN{0730-0301}
\urldef\tempurl%
\url{https://doi.org/10.1145/1531326.1531330}
\showDOI{\tempurl}


\bibitem[Barron et~al\mbox{.}(2022)]%
        {barron2022mip360}
\bibfield{author}{\bibinfo{person}{Jonathan~T Barron}, \bibinfo{person}{Ben Mildenhall}, \bibinfo{person}{Dor Verbin}, \bibinfo{person}{Pratul~P Srinivasan}, {and} \bibinfo{person}{Peter Hedman}.} \bibinfo{year}{2022}\natexlab{}.
\newblock \showarticletitle{Mip-nerf 360: Unbounded anti-aliased neural radiance fields}. In \bibinfo{booktitle}{\emph{Proceedings of the IEEE/CVF Conference on Computer Vision and Pattern Recognition}}. \bibinfo{pages}{5470--5479}.
\newblock


\bibitem[Cen et~al\mbox{.}(2023)]%
        {cen2023sa3d}
\bibfield{author}{\bibinfo{person}{Jiazhong Cen}, \bibinfo{person}{Zanwei Zhou}, \bibinfo{person}{Jiemin Fang}, \bibinfo{person}{Chen Yang}, \bibinfo{person}{Wei Shen}, \bibinfo{person}{Lingxi Xie}, \bibinfo{person}{Dongsheng Jiang}, \bibinfo{person}{Xiaopeng Zhang}, {and} \bibinfo{person}{Qi Tian}.} \bibinfo{year}{2023}\natexlab{}.
\newblock \showarticletitle{Segment Anything in 3D with NeRFs}. In \bibinfo{booktitle}{\emph{NeurIPS}}.
\newblock


\bibitem[Chen et~al\mbox{.}(2023a)]%
        {chen2023gaussianeditor}
\bibfield{author}{\bibinfo{person}{Yiwen Chen}, \bibinfo{person}{Zilong Chen}, \bibinfo{person}{Chi Zhang}, \bibinfo{person}{Feng Wang}, \bibinfo{person}{Xiaofeng Yang}, \bibinfo{person}{Yikai Wang}, \bibinfo{person}{Zhongang Cai}, \bibinfo{person}{Lei Yang}, \bibinfo{person}{Huaping Liu}, {and} \bibinfo{person}{Guosheng Lin}.} \bibinfo{year}{2023}\natexlab{a}.
\newblock \bibinfo{title}{GaussianEditor: Swift and Controllable 3D Editing with Gaussian Splatting}.
\newblock
\newblock
\showeprint[arxiv]{2311.14521}~[cs.CV]


\bibitem[Chen et~al\mbox{.}(2023b)]%
        {chen2023mobilenerf}
\bibfield{author}{\bibinfo{person}{Zhiqin Chen}, \bibinfo{person}{Thomas Funkhouser}, \bibinfo{person}{Peter Hedman}, {and} \bibinfo{person}{Andrea Tagliasacchi}.} \bibinfo{year}{2023}\natexlab{b}.
\newblock \showarticletitle{Mobilenerf: Exploiting the polygon rasterization pipeline for efficient neural field rendering on mobile architectures}. In \bibinfo{booktitle}{\emph{Proceedings of the IEEE/CVF Conference on Computer Vision and Pattern Recognition}}. \bibinfo{pages}{16569--16578}.
\newblock


\bibitem[Darabi et~al\mbox{.}(2012)]%
        {darabi2012imagemeld}
\bibfield{author}{\bibinfo{person}{Soheil Darabi}, \bibinfo{person}{Eli Shechtman}, \bibinfo{person}{Connelly Barnes}, \bibinfo{person}{Dan~B Goldman}, {and} \bibinfo{person}{Pradeep Sen}.} \bibinfo{year}{2012}\natexlab{}.
\newblock \showarticletitle{Image melding: Combining inconsistent images using patch-based synthesis}.
\newblock \bibinfo{journal}{\emph{ACM Transactions on graphics (TOG)}} \bibinfo{volume}{31}, \bibinfo{number}{4} (\bibinfo{year}{2012}), \bibinfo{pages}{1--10}.
\newblock


\bibitem[Dessein et~al\mbox{.}(2014)]%
        {dessein2014seamless}
\bibfield{author}{\bibinfo{person}{Arnaud Dessein}, \bibinfo{person}{William~AP Smith}, \bibinfo{person}{Richard~C Wilson}, {and} \bibinfo{person}{Edwin~R Hancock}.} \bibinfo{year}{2014}\natexlab{}.
\newblock \showarticletitle{Seamless texture stitching on a 3D mesh by poisson blending in patches}. In \bibinfo{booktitle}{\emph{2014 IEEE International Conference on Image Processing (ICIP)}}. IEEE, \bibinfo{pages}{2031--2035}.
\newblock


\bibitem[Efros and Leung(1999)]%
        {efros1999texture}
\bibfield{author}{\bibinfo{person}{Alexei~A Efros} {and} \bibinfo{person}{Thomas~K Leung}.} \bibinfo{year}{1999}\natexlab{}.
\newblock \showarticletitle{Texture synthesis by non-parametric sampling}. In \bibinfo{booktitle}{\emph{Proceedings of the seventh IEEE international conference on computer vision}}, Vol.~\bibinfo{volume}{2}. IEEE, \bibinfo{pages}{1033--1038}.
\newblock


\bibitem[Fisher et~al\mbox{.}(2012)]%
        {fisher2012example}
\bibfield{author}{\bibinfo{person}{Matthew Fisher}, \bibinfo{person}{Daniel Ritchie}, \bibinfo{person}{Manolis Savva}, \bibinfo{person}{Thomas Funkhouser}, {and} \bibinfo{person}{Pat Hanrahan}.} \bibinfo{year}{2012}\natexlab{}.
\newblock \showarticletitle{Example-based synthesis of 3D object arrangements}.
\newblock \bibinfo{journal}{\emph{ACM Transactions on Graphics (TOG)}} \bibinfo{volume}{31}, \bibinfo{number}{6} (\bibinfo{year}{2012}), \bibinfo{pages}{1--11}.
\newblock


\bibitem[Fu et~al\mbox{.}(2022)]%
        {fu2022panoptic}
\bibfield{author}{\bibinfo{person}{Xiao Fu}, \bibinfo{person}{Shangzhan Zhang}, \bibinfo{person}{Tianrun Chen}, \bibinfo{person}{Yichong Lu}, \bibinfo{person}{Lanyun Zhu}, \bibinfo{person}{Xiaowei Zhou}, \bibinfo{person}{Andreas Geiger}, {and} \bibinfo{person}{Yiyi Liao}.} \bibinfo{year}{2022}\natexlab{}.
\newblock \showarticletitle{Panoptic nerf: 3d-to-2d label transfer for panoptic urban scene segmentation}. In \bibinfo{booktitle}{\emph{2022 International Conference on 3D Vision (3DV)}}. IEEE, \bibinfo{pages}{1--11}.
\newblock


\bibitem[Funkhouser et~al\mbox{.}(2004)]%
        {funkhouser2004modeling}
\bibfield{author}{\bibinfo{person}{Thomas Funkhouser}, \bibinfo{person}{Michael Kazhdan}, \bibinfo{person}{Philip Shilane}, \bibinfo{person}{Patrick Min}, \bibinfo{person}{William Kiefer}, \bibinfo{person}{Ayellet Tal}, \bibinfo{person}{Szymon Rusinkiewicz}, {and} \bibinfo{person}{David Dobkin}.} \bibinfo{year}{2004}\natexlab{}.
\newblock \showarticletitle{Modeling by example}.
\newblock \bibinfo{journal}{\emph{ACM transactions on graphics (TOG)}} \bibinfo{volume}{23}, \bibinfo{number}{3} (\bibinfo{year}{2004}), \bibinfo{pages}{652--663}.
\newblock


\bibitem[Gao et~al\mbox{.}(2023)]%
        {Relight3DG2023}
\bibfield{author}{\bibinfo{person}{Jian Gao}, \bibinfo{person}{Chun Gu}, \bibinfo{person}{Youtian Lin}, \bibinfo{person}{Hao Zhu}, \bibinfo{person}{Xun Cao}, \bibinfo{person}{Li Zhang}, {and} \bibinfo{person}{Yao Yao}.} \bibinfo{year}{2023}\natexlab{}.
\newblock \showarticletitle{Relightable 3D Gaussian: Real-time Point Cloud Relighting with BRDF Decomposition and Ray Tracing}.
\newblock \bibinfo{journal}{\emph{arXiv:2311.16043}} (\bibinfo{year}{2023}).
\newblock


\bibitem[Gao et~al\mbox{.}(2024)]%
        {gao2023aaaiAGIF}
\bibfield{author}{\bibinfo{person}{Xinyu Gao}, \bibinfo{person}{Ziyi Yang}, \bibinfo{person}{Yunlu Zhao}, \bibinfo{person}{Yuxiang Sun}, \bibinfo{person}{Xiaogang Jin}, {and} \bibinfo{person}{Changqing Zou}.} \bibinfo{year}{2024}\natexlab{}.
\newblock \showarticletitle{A General Implicit Framework for Fast NeRF Composition and Rendering}. In \bibinfo{booktitle}{\emph{Proceedings of the AAAI Conference on Artificial Intelligence}}, Vol.~\bibinfo{volume}{38}. \bibinfo{pages}{1833--1841}.
\newblock


\bibitem[Gong et~al\mbox{.}(2023)]%
        {gong2023seamlessnerf}
\bibfield{author}{\bibinfo{person}{Bingchen Gong}, \bibinfo{person}{Yuehao Wang}, \bibinfo{person}{Xiaoguang Han}, {and} \bibinfo{person}{Qi Dou}.} \bibinfo{year}{2023}\natexlab{}.
\newblock \showarticletitle{SeamlessNeRF: Stitching Part NeRFs with Gradient Propagation}. In \bibinfo{booktitle}{\emph{SIGGRAPH Asia 2023 Conference Papers}}. \bibinfo{pages}{1--10}.
\newblock


\bibitem[Huang et~al\mbox{.}(2024)]%
        {huang2024sc}
\bibfield{author}{\bibinfo{person}{Yi-Hua Huang}, \bibinfo{person}{Yang-Tian Sun}, \bibinfo{person}{Ziyi Yang}, \bibinfo{person}{Xiaoyang Lyu}, \bibinfo{person}{Yan-Pei Cao}, {and} \bibinfo{person}{Xiaojuan Qi}.} \bibinfo{year}{2024}\natexlab{}.
\newblock \showarticletitle{Sc-gs: Sparse-controlled gaussian splatting for editable dynamic scenes}. In \bibinfo{booktitle}{\emph{Proceedings of the IEEE/CVF Conference on Computer Vision and Pattern Recognition}}. \bibinfo{pages}{4220--4230}.
\newblock


\bibitem[Igarashi et~al\mbox{.}(2005)]%
        {igarashi2005rigid}
\bibfield{author}{\bibinfo{person}{Takeo Igarashi}, \bibinfo{person}{Tomer Moscovich}, {and} \bibinfo{person}{John~F Hughes}.} \bibinfo{year}{2005}\natexlab{}.
\newblock \showarticletitle{As-rigid-as-possible shape manipulation}.
\newblock \bibinfo{journal}{\emph{ACM transactions on Graphics (TOG)}} \bibinfo{volume}{24}, \bibinfo{number}{3} (\bibinfo{year}{2005}), \bibinfo{pages}{1134--1141}.
\newblock


\bibitem[Ivanic and Ruedenberg(1996)]%
        {ivanic1996shrotation}
\bibfield{author}{\bibinfo{person}{Joseph Ivanic} {and} \bibinfo{person}{Klaus Ruedenberg}.} \bibinfo{year}{1996}\natexlab{}.
\newblock \showarticletitle{Rotation matrices for real spherical harmonics. Direct determination by recursion}.
\newblock \bibinfo{journal}{\emph{The Journal of Physical Chemistry}} \bibinfo{volume}{100}, \bibinfo{number}{15} (\bibinfo{year}{1996}), \bibinfo{pages}{6342--6347}.
\newblock


\bibitem[Katz et~al\mbox{.}(2005)]%
        {katz2005mesh}
\bibfield{author}{\bibinfo{person}{Sagi Katz}, \bibinfo{person}{George Leifman}, {and} \bibinfo{person}{Ayellet Tal}.} \bibinfo{year}{2005}\natexlab{}.
\newblock \showarticletitle{Mesh segmentation using feature point and core extraction}.
\newblock \bibinfo{journal}{\emph{The Visual Computer}}  \bibinfo{volume}{21} (\bibinfo{year}{2005}), \bibinfo{pages}{649--658}.
\newblock


\bibitem[Kerbl et~al\mbox{.}(2023)]%
        {kerbl3Dgaussians}
\bibfield{author}{\bibinfo{person}{Bernhard Kerbl}, \bibinfo{person}{Georgios Kopanas}, \bibinfo{person}{Thomas Leimk{\"u}hler}, {and} \bibinfo{person}{George Drettakis}.} \bibinfo{year}{2023}\natexlab{}.
\newblock \showarticletitle{3D Gaussian Splatting for Real-Time Radiance Field Rendering}.
\newblock \bibinfo{journal}{\emph{ACM Transactions on Graphics}} \bibinfo{volume}{42}, \bibinfo{number}{4} (\bibinfo{date}{July} \bibinfo{year}{2023}).
\newblock
\urldef\tempurl%
\url{https://repo-sam.inria.fr/fungraph/3d-gaussian-splatting/}
\showURL{%
\tempurl}


\bibitem[Kerr et~al\mbox{.}(2023)]%
        {kerr2023lerf}
\bibfield{author}{\bibinfo{person}{Justin Kerr}, \bibinfo{person}{Chung~Min Kim}, \bibinfo{person}{Ken Goldberg}, \bibinfo{person}{Angjoo Kanazawa}, {and} \bibinfo{person}{Matthew Tancik}.} \bibinfo{year}{2023}\natexlab{}.
\newblock \showarticletitle{Lerf: Language embedded radiance fields}. In \bibinfo{booktitle}{\emph{Proceedings of the IEEE/CVF International Conference on Computer Vision}}. \bibinfo{pages}{19729--19739}.
\newblock


\bibitem[Kirillov et~al\mbox{.}(2023)]%
        {kirillov2023segany}
\bibfield{author}{\bibinfo{person}{Alexander Kirillov}, \bibinfo{person}{Eric Mintun}, \bibinfo{person}{Nikhila Ravi}, \bibinfo{person}{Hanzi Mao}, \bibinfo{person}{Chloe Rolland}, \bibinfo{person}{Laura Gustafson}, \bibinfo{person}{Tete Xiao}, \bibinfo{person}{Spencer Whitehead}, \bibinfo{person}{Alexander~C. Berg}, \bibinfo{person}{Wan-Yen Lo}, \bibinfo{person}{Piotr Doll{\'a}r}, {and} \bibinfo{person}{Ross Girshick}.} \bibinfo{year}{2023}\natexlab{}.
\newblock \showarticletitle{Segment Anything}.
\newblock \bibinfo{journal}{\emph{arXiv:2304.02643}} (\bibinfo{year}{2023}).
\newblock


\bibitem[Kreavoy et~al\mbox{.}(2007)]%
        {kreavoy2007model}
\bibfield{author}{\bibinfo{person}{Vladislav Kreavoy}, \bibinfo{person}{Dan Julius}, {and} \bibinfo{person}{Alla Sheffer}.} \bibinfo{year}{2007}\natexlab{}.
\newblock \showarticletitle{Model composition from interchangeable components}. In \bibinfo{booktitle}{\emph{15th Pacific Conference on Computer Graphics and Applications (PG'07)}}. IEEE, \bibinfo{pages}{129--138}.
\newblock


\bibitem[Kundu et~al\mbox{.}(2022)]%
        {kundu2022panoptic}
\bibfield{author}{\bibinfo{person}{Abhijit Kundu}, \bibinfo{person}{Kyle Genova}, \bibinfo{person}{Xiaoqi Yin}, \bibinfo{person}{Alireza Fathi}, \bibinfo{person}{Caroline Pantofaru}, \bibinfo{person}{Leonidas~J Guibas}, \bibinfo{person}{Andrea Tagliasacchi}, \bibinfo{person}{Frank Dellaert}, {and} \bibinfo{person}{Thomas Funkhouser}.} \bibinfo{year}{2022}\natexlab{}.
\newblock \showarticletitle{Panoptic neural fields: A semantic object-aware neural scene representation}. In \bibinfo{booktitle}{\emph{Proceedings of the IEEE/CVF Conference on Computer Vision and Pattern Recognition}}. \bibinfo{pages}{12871--12881}.
\newblock


\bibitem[Kwatra et~al\mbox{.}(2005)]%
        {Kwatra2005TextureOF}
\bibfield{author}{\bibinfo{person}{Vivek Kwatra}, \bibinfo{person}{Irfan Essa}, \bibinfo{person}{Aaron Bobick}, {and} \bibinfo{person}{Nipun Kwatra}.} \bibinfo{year}{2005}\natexlab{}.
\newblock \showarticletitle{Texture Optimization for Example-Based Synthesis}.
\newblock \bibinfo{journal}{\emph{ACM Trans. Graph.}} \bibinfo{volume}{24}, \bibinfo{number}{3} (\bibinfo{date}{jul} \bibinfo{year}{2005}), \bibinfo{pages}{795–802}.
\newblock
\showISSN{0730-0301}
\urldef\tempurl%
\url{https://doi.org/10.1145/1073204.1073263}
\showDOI{\tempurl}


\bibitem[Li et~al\mbox{.}(2023b)]%
        {li2023compressing}
\bibfield{author}{\bibinfo{person}{Lingzhi Li}, \bibinfo{person}{Zhen Shen}, \bibinfo{person}{Zhongshu Wang}, \bibinfo{person}{Li Shen}, {and} \bibinfo{person}{Liefeng Bo}.} \bibinfo{year}{2023}\natexlab{b}.
\newblock \showarticletitle{Compressing volumetric radiance fields to 1 mb}. In \bibinfo{booktitle}{\emph{Proceedings of the IEEE/CVF Conference on Computer Vision and Pattern Recognition}}. \bibinfo{pages}{4222--4231}.
\newblock


\bibitem[Li et~al\mbox{.}(2023a)]%
        {li2023patchexample}
\bibfield{author}{\bibinfo{person}{Weiyu Li}, \bibinfo{person}{Xuelin Chen}, \bibinfo{person}{Jue Wang}, {and} \bibinfo{person}{Baoquan Chen}.} \bibinfo{year}{2023}\natexlab{a}.
\newblock \showarticletitle{Patch-based 3D Natural Scene Generation from a Single Example}. In \bibinfo{booktitle}{\emph{Proceedings of the IEEE/CVF Conference on Computer Vision and Pattern Recognition}}. \bibinfo{pages}{16762--16772}.
\newblock


\bibitem[Li et~al\mbox{.}(2024)]%
        {li2024spacetime}
\bibfield{author}{\bibinfo{person}{Zhan Li}, \bibinfo{person}{Zhang Chen}, \bibinfo{person}{Zhong Li}, {and} \bibinfo{person}{Yi Xu}.} \bibinfo{year}{2024}\natexlab{}.
\newblock \showarticletitle{Spacetime gaussian feature splatting for real-time dynamic view synthesis}. In \bibinfo{booktitle}{\emph{Proceedings of the IEEE/CVF Conference on Computer Vision and Pattern Recognition}}. \bibinfo{pages}{8508--8520}.
\newblock


\bibitem[Liang et~al\mbox{.}(2023)]%
        {EnVision2023luciddreamer}
\bibfield{author}{\bibinfo{person}{Yixun Liang}, \bibinfo{person}{Xin Yang}, \bibinfo{person}{Jiantao Lin}, \bibinfo{person}{Haodong Li}, \bibinfo{person}{Xiaogang Xu}, {and} \bibinfo{person}{Yingcong Chen}.} \bibinfo{year}{2023}\natexlab{}.
\newblock \bibinfo{title}{LucidDreamer: Towards High-Fidelity Text-to-3D Generation via Interval Score Matching}.
\newblock
\newblock
\showeprint[arxiv]{2311.11284}~[cs.CV]


\bibitem[Lin et~al\mbox{.}(2022)]%
        {lin2022enerf}
\bibfield{author}{\bibinfo{person}{Haotong Lin}, \bibinfo{person}{Sida Peng}, \bibinfo{person}{Zhen Xu}, \bibinfo{person}{Yunzhi Yan}, \bibinfo{person}{Qing Shuai}, \bibinfo{person}{Hujun Bao}, {and} \bibinfo{person}{Xiaowei Zhou}.} \bibinfo{year}{2022}\natexlab{}.
\newblock \showarticletitle{Efficient Neural Radiance Fields for Interactive Free-viewpoint Video}. In \bibinfo{booktitle}{\emph{SIGGRAPH Asia Conference Proceedings}}.
\newblock


\bibitem[Liu et~al\mbox{.}(2023b)]%
        {liu2023neuralimposter}
\bibfield{author}{\bibinfo{person}{Ruiyang Liu}, \bibinfo{person}{Jinxu Xiang}, \bibinfo{person}{Bowen Zhao}, \bibinfo{person}{Ran Zhang}, \bibinfo{person}{Jingyi Yu}, {and} \bibinfo{person}{Changxi Zheng}.} \bibinfo{year}{2023}\natexlab{b}.
\newblock \showarticletitle{Neural impostor: Editing neural radiance fields with explicit shape manipulation}. In \bibinfo{booktitle}{\emph{Computer Graphics Forum}}. Wiley Online Library, \bibinfo{pages}{e14981}.
\newblock


\bibitem[Liu et~al\mbox{.}(2023a)]%
        {liu2023nero}
\bibfield{author}{\bibinfo{person}{Yuan Liu}, \bibinfo{person}{Peng Wang}, \bibinfo{person}{Cheng Lin}, \bibinfo{person}{Xiaoxiao Long}, \bibinfo{person}{Jiepeng Wang}, \bibinfo{person}{Lingjie Liu}, \bibinfo{person}{Taku Komura}, {and} \bibinfo{person}{Wenping Wang}.} \bibinfo{year}{2023}\natexlab{a}.
\newblock \showarticletitle{NeRO: Neural Geometry and BRDF Reconstruction of Reflective Objects from Multiview Images}.
\newblock \bibinfo{journal}{\emph{ACM Trans. Graph.}} \bibinfo{volume}{42}, \bibinfo{number}{4}, Article \bibinfo{articleno}{114} (\bibinfo{date}{jul} \bibinfo{year}{2023}), \bibinfo{numpages}{22}~pages.
\newblock
\showISSN{0730-0301}
\urldef\tempurl%
\url{https://doi.org/10.1145/3592134}
\showDOI{\tempurl}


\bibitem[Merrell(2007)]%
        {merrell2007example}
\bibfield{author}{\bibinfo{person}{Paul Merrell}.} \bibinfo{year}{2007}\natexlab{}.
\newblock \showarticletitle{Example-based model synthesis}. In \bibinfo{booktitle}{\emph{Proceedings of the 2007 symposium on Interactive 3D graphics and games}}. \bibinfo{pages}{105--112}.
\newblock


\bibitem[Mirzaei et~al\mbox{.}(2023)]%
        {mirzaei2023spin}
\bibfield{author}{\bibinfo{person}{Ashkan Mirzaei}, \bibinfo{person}{Tristan Aumentado-Armstrong}, \bibinfo{person}{Konstantinos~G Derpanis}, \bibinfo{person}{Jonathan Kelly}, \bibinfo{person}{Marcus~A Brubaker}, \bibinfo{person}{Igor Gilitschenski}, {and} \bibinfo{person}{Alex Levinshtein}.} \bibinfo{year}{2023}\natexlab{}.
\newblock \showarticletitle{SPIn-NeRF: Multiview segmentation and perceptual inpainting with neural radiance fields}. In \bibinfo{booktitle}{\emph{Proceedings of the IEEE/CVF Conference on Computer Vision and Pattern Recognition}}. \bibinfo{pages}{20669--20679}.
\newblock


\bibitem[Nguyen-Phuoc et~al\mbox{.}(2022)]%
        {meta2022snerf}
\bibfield{author}{\bibinfo{person}{Thu Nguyen-Phuoc}, \bibinfo{person}{Feng Liu}, {and} \bibinfo{person}{Lei Xiao}.} \bibinfo{year}{2022}\natexlab{}.
\newblock \showarticletitle{SNeRF: Stylized Neural Implicit Representations for 3D Scenes}.
\newblock  \bibinfo{volume}{41}, \bibinfo{number}{4}, Article \bibinfo{articleno}{142} (\bibinfo{date}{jul} \bibinfo{year}{2022}), \bibinfo{numpages}{11}~pages.
\newblock
\showISSN{0730-0301}
\urldef\tempurl%
\url{https://doi.org/10.1145/3528223.3530107}
\showDOI{\tempurl}


\bibitem[Ost et~al\mbox{.}(2021)]%
        {ost2021neural}
\bibfield{author}{\bibinfo{person}{Julian Ost}, \bibinfo{person}{Fahim Mannan}, \bibinfo{person}{Nils Thuerey}, \bibinfo{person}{Julian Knodt}, {and} \bibinfo{person}{Felix Heide}.} \bibinfo{year}{2021}\natexlab{}.
\newblock \showarticletitle{Neural scene graphs for dynamic scenes}. In \bibinfo{booktitle}{\emph{Proceedings of the IEEE/CVF Conference on Computer Vision and Pattern Recognition}}. \bibinfo{pages}{2856--2865}.
\newblock


\bibitem[P{\'e}rez et~al\mbox{.}(2023)]%
        {perez2023poisson}
\bibfield{author}{\bibinfo{person}{Patrick P{\'e}rez}, \bibinfo{person}{Michel Gangnet}, {and} \bibinfo{person}{Andrew Blake}.} \bibinfo{year}{2023}\natexlab{}.
\newblock \showarticletitle{Poisson image editing}.
\newblock In \bibinfo{booktitle}{\emph{Seminal Graphics Papers: Pushing the Boundaries, Volume 2}}. \bibinfo{pages}{577--582}.
\newblock


\bibitem[Po and Wetzstein(2023)]%
        {po2023compositionaldiff}
\bibfield{author}{\bibinfo{person}{Ryan Po} {and} \bibinfo{person}{Gordon Wetzstein}.} \bibinfo{year}{2023}\natexlab{}.
\newblock \showarticletitle{Compositional 3d scene generation using locally conditioned diffusion}.
\newblock \bibinfo{journal}{\emph{arXiv preprint arXiv:2303.12218}} (\bibinfo{year}{2023}).
\newblock


\bibitem[Qiao et~al\mbox{.}(2023)]%
        {qiao2023dynamic}
\bibfield{author}{\bibinfo{person}{Yi-Ling Qiao}, \bibinfo{person}{Alexander Gao}, \bibinfo{person}{Yiran Xu}, \bibinfo{person}{Yue Feng}, \bibinfo{person}{Jia-Bin Huang}, {and} \bibinfo{person}{Ming~C Lin}.} \bibinfo{year}{2023}\natexlab{}.
\newblock \showarticletitle{Dynamic mesh-aware radiance fields}. In \bibinfo{booktitle}{\emph{Proceedings of the IEEE/CVF International Conference on Computer Vision}}. \bibinfo{pages}{385--396}.
\newblock


\bibitem[Rocchini et~al\mbox{.}(1999)]%
        {rocchini1999multiple}
\bibfield{author}{\bibinfo{person}{Claudio Rocchini}, \bibinfo{person}{Paolo Cignoni}, \bibinfo{person}{Claudio Montani}, {and} \bibinfo{person}{Roberto Scopigno}.} \bibinfo{year}{1999}\natexlab{}.
\newblock \showarticletitle{Multiple textures stitching and blending on 3D objects}. In \bibinfo{booktitle}{\emph{Rendering Techniques’ 99: Proceedings of the Eurographics Workshop in Granada, Spain, June 21--23, 1999 10}}. Springer, \bibinfo{pages}{119--130}.
\newblock


\bibitem[Shuai et~al\mbox{.}(2022)]%
        {shuai2022multinb}
\bibfield{author}{\bibinfo{person}{Qing Shuai}, \bibinfo{person}{Chen Geng}, \bibinfo{person}{Qi Fang}, \bibinfo{person}{Sida Peng}, \bibinfo{person}{Wenhao Shen}, \bibinfo{person}{Xiaowei Zhou}, {and} \bibinfo{person}{Hujun Bao}.} \bibinfo{year}{2022}\natexlab{}.
\newblock \showarticletitle{Novel view synthesis of human interactions from sparse multi-view videos}. In \bibinfo{booktitle}{\emph{SIGGRAPH Conference Proceedings}}. \bibinfo{pages}{1--10}.
\newblock


\bibitem[Sobel et~al\mbox{.}(1968)]%
        {sobel19683x3}
\bibfield{author}{\bibinfo{person}{Irwin Sobel}, \bibinfo{person}{Gary Feldman}, {et~al\mbox{.}}} \bibinfo{year}{1968}\natexlab{}.
\newblock \showarticletitle{A 3x3 isotropic gradient operator for image processing}.
\newblock \bibinfo{journal}{\emph{a talk at the Stanford Artificial Project in}} (\bibinfo{year}{1968}), \bibinfo{pages}{271--272}.
\newblock


\bibitem[Tancik et~al\mbox{.}(2022)]%
        {tancik2022block}
\bibfield{author}{\bibinfo{person}{Matthew Tancik}, \bibinfo{person}{Vincent Casser}, \bibinfo{person}{Xinchen Yan}, \bibinfo{person}{Sabeek Pradhan}, \bibinfo{person}{Ben Mildenhall}, \bibinfo{person}{Pratul~P Srinivasan}, \bibinfo{person}{Jonathan~T Barron}, {and} \bibinfo{person}{Henrik Kretzschmar}.} \bibinfo{year}{2022}\natexlab{}.
\newblock \showarticletitle{Block-nerf: Scalable large scene neural view synthesis}. In \bibinfo{booktitle}{\emph{Proceedings of the IEEE/CVF Conference on Computer Vision and Pattern Recognition}}. \bibinfo{pages}{8248--8258}.
\newblock


\bibitem[Tang et~al\mbox{.}(2023a)]%
        {tang2023dreamgaussian}
\bibfield{author}{\bibinfo{person}{Jiaxiang Tang}, \bibinfo{person}{Jiawei Ren}, \bibinfo{person}{Hang Zhou}, \bibinfo{person}{Ziwei Liu}, {and} \bibinfo{person}{Gang Zeng}.} \bibinfo{year}{2023}\natexlab{a}.
\newblock \showarticletitle{DreamGaussian: Generative Gaussian Splatting for Efficient 3D Content Creation}.
\newblock \bibinfo{journal}{\emph{arXiv preprint arXiv:2309.16653}} (\bibinfo{year}{2023}).
\newblock


\bibitem[Tang et~al\mbox{.}(2023b)]%
        {Tang_2023_ICCV}
\bibfield{author}{\bibinfo{person}{Jiaxiang Tang}, \bibinfo{person}{Hang Zhou}, \bibinfo{person}{Xiaokang Chen}, \bibinfo{person}{Tianshu Hu}, \bibinfo{person}{Errui Ding}, \bibinfo{person}{Jingdong Wang}, {and} \bibinfo{person}{Gang Zeng}.} \bibinfo{year}{2023}\natexlab{b}.
\newblock \showarticletitle{Delicate Textured Mesh Recovery from NeRF via Adaptive Surface Refinement}. In \bibinfo{booktitle}{\emph{Proceedings of the IEEE/CVF International Conference on Computer Vision (ICCV)}}. \bibinfo{pages}{17739--17749}.
\newblock


\bibitem[Wang et~al\mbox{.}(2023)]%
        {Wang_2023_CVPR}
\bibfield{author}{\bibinfo{person}{Liao Wang}, \bibinfo{person}{Qiang Hu}, \bibinfo{person}{Qihan He}, \bibinfo{person}{Ziyu Wang}, \bibinfo{person}{Jingyi Yu}, \bibinfo{person}{Tinne Tuytelaars}, \bibinfo{person}{Lan Xu}, {and} \bibinfo{person}{Minye Wu}.} \bibinfo{year}{2023}\natexlab{}.
\newblock \showarticletitle{Neural Residual Radiance Fields for Streamably Free-Viewpoint Videos}. In \bibinfo{booktitle}{\emph{Proceedings of the IEEE/CVF Conference on Computer Vision and Pattern Recognition (CVPR)}}. \bibinfo{pages}{76--87}.
\newblock


\bibitem[Wei et~al\mbox{.}(2009)]%
        {wei2009state}
\bibfield{author}{\bibinfo{person}{Li-Yi Wei}, \bibinfo{person}{Sylvain Lefebvre}, \bibinfo{person}{Vivek Kwatra}, {and} \bibinfo{person}{Greg Turk}.} \bibinfo{year}{2009}\natexlab{}.
\newblock \showarticletitle{State of the art in example-based texture synthesis}.
\newblock \bibinfo{journal}{\emph{Eurographics 2009, State of the Art Report, EG-STAR}} (\bibinfo{year}{2009}), \bibinfo{pages}{93--117}.
\newblock


\bibitem[Wu et~al\mbox{.}(2023)]%
        {wu2023dover}
\bibfield{author}{\bibinfo{person}{Haoning Wu}, \bibinfo{person}{Erli Zhang}, \bibinfo{person}{Liang Liao}, \bibinfo{person}{Chaofeng Chen}, \bibinfo{person}{Jingwen~Hou Hou}, \bibinfo{person}{Annan Wang}, \bibinfo{person}{Wenxiu~Sun Sun}, \bibinfo{person}{Qiong Yan}, {and} \bibinfo{person}{Weisi Lin}.} \bibinfo{year}{2023}\natexlab{}.
\newblock \showarticletitle{Exploring Video Quality Assessment on User Generated Contents from Aesthetic and Technical Perspectives}. In \bibinfo{booktitle}{\emph{International Conference on Computer Vision (ICCV)}}.
\newblock


\bibitem[Wu et~al\mbox{.}(2022)]%
        {wu2022object}
\bibfield{author}{\bibinfo{person}{Qianyi Wu}, \bibinfo{person}{Xian Liu}, \bibinfo{person}{Yuedong Chen}, \bibinfo{person}{Kejie Li}, \bibinfo{person}{Chuanxia Zheng}, \bibinfo{person}{Jianfei Cai}, {and} \bibinfo{person}{Jianmin Zheng}.} \bibinfo{year}{2022}\natexlab{}.
\newblock \showarticletitle{Object-compositional neural implicit surfaces}. In \bibinfo{booktitle}{\emph{Computer Vision--ECCV 2022: 17th European Conference, Tel Aviv, Israel, October 23--27, 2022, Proceedings, Part XXVII}}. Springer, \bibinfo{pages}{197--213}.
\newblock


\bibitem[Wu and Zheng(2022)]%
        {wu2022togexample}
\bibfield{author}{\bibinfo{person}{Rundi Wu} {and} \bibinfo{person}{Changxi Zheng}.} \bibinfo{year}{2022}\natexlab{}.
\newblock \showarticletitle{Learning to Generate 3D Shapes from a Single Example}.
\newblock \bibinfo{journal}{\emph{ACM Trans. Graph.}} \bibinfo{volume}{41}, \bibinfo{number}{6}, Article \bibinfo{articleno}{224} (\bibinfo{date}{nov} \bibinfo{year}{2022}), \bibinfo{numpages}{19}~pages.
\newblock
\showISSN{0730-0301}
\urldef\tempurl%
\url{https://doi.org/10.1145/3550454.3555480}
\showDOI{\tempurl}


\bibitem[Yang et~al\mbox{.}(2022)]%
        {yang2022neumesh}
\bibfield{author}{\bibinfo{person}{Bangbang Yang}, \bibinfo{person}{Chong Bao}, \bibinfo{person}{Junyi Zeng}, \bibinfo{person}{Hujun Bao}, \bibinfo{person}{Yinda Zhang}, \bibinfo{person}{Zhaopeng Cui}, {and} \bibinfo{person}{Guofeng Zhang}.} \bibinfo{year}{2022}\natexlab{}.
\newblock \showarticletitle{Neumesh: Learning disentangled neural mesh-based implicit field for geometry and texture editing}. In \bibinfo{booktitle}{\emph{European Conference on Computer Vision}}. Springer, \bibinfo{pages}{597--614}.
\newblock


\bibitem[Yang et~al\mbox{.}(2021)]%
        {yang2021learning}
\bibfield{author}{\bibinfo{person}{Bangbang Yang}, \bibinfo{person}{Yinda Zhang}, \bibinfo{person}{Yinghao Xu}, \bibinfo{person}{Yijin Li}, \bibinfo{person}{Han Zhou}, \bibinfo{person}{Hujun Bao}, \bibinfo{person}{Guofeng Zhang}, {and} \bibinfo{person}{Zhaopeng Cui}.} \bibinfo{year}{2021}\natexlab{}.
\newblock \showarticletitle{Learning object-compositional neural radiance field for editable scene rendering}. In \bibinfo{booktitle}{\emph{Proceedings of the IEEE/CVF International Conference on Computer Vision}}. \bibinfo{pages}{13779--13788}.
\newblock


\bibitem[Yang et~al\mbox{.}(2023)]%
        {yang2023unisim}
\bibfield{author}{\bibinfo{person}{Ze Yang}, \bibinfo{person}{Yun Chen}, \bibinfo{person}{Jingkang Wang}, \bibinfo{person}{Sivabalan Manivasagam}, \bibinfo{person}{Wei-Chiu Ma}, \bibinfo{person}{Anqi~Joyce Yang}, {and} \bibinfo{person}{Raquel Urtasun}.} \bibinfo{year}{2023}\natexlab{}.
\newblock \showarticletitle{UniSim: A Neural Closed-Loop Sensor Simulator}. In \bibinfo{booktitle}{\emph{Proceedings of the IEEE/CVF Conference on Computer Vision and Pattern Recognition}}. \bibinfo{pages}{1389--1399}.
\newblock


\bibitem[Yang et~al\mbox{.}(2024a)]%
        {yang2024deformable}
\bibfield{author}{\bibinfo{person}{Ziyi Yang}, \bibinfo{person}{Xinyu Gao}, \bibinfo{person}{Wen Zhou}, \bibinfo{person}{Shaohui Jiao}, \bibinfo{person}{Yuqing Zhang}, {and} \bibinfo{person}{Xiaogang Jin}.} \bibinfo{year}{2024}\natexlab{a}.
\newblock \showarticletitle{Deformable 3d gaussians for high-fidelity monocular dynamic scene reconstruction}. In \bibinfo{booktitle}{\emph{Proceedings of the IEEE/CVF Conference on Computer Vision and Pattern Recognition}}. \bibinfo{pages}{20331--20341}.
\newblock


\bibitem[Yang et~al\mbox{.}(2024b)]%
        {yang2023gs4d}
\bibfield{author}{\bibinfo{person}{Zeyu Yang}, \bibinfo{person}{Hongye Yang}, \bibinfo{person}{Zijie Pan}, {and} \bibinfo{person}{Li Zhang}.} \bibinfo{year}{2024}\natexlab{b}.
\newblock \showarticletitle{Real-time Photorealistic Dynamic Scene Representation and Rendering with 4D Gaussian Splatting}. In \bibinfo{booktitle}{\emph{International Conference on Learning Representations (ICLR)}}.
\newblock


\bibitem[Yao et~al\mbox{.}(2020)]%
        {blendedMVS}
\bibfield{author}{\bibinfo{person}{Yao Yao}, \bibinfo{person}{Zixin Luo}, \bibinfo{person}{Shiwei Li}, \bibinfo{person}{Jingyang Zhang}, \bibinfo{person}{Yufan Ren}, \bibinfo{person}{Lei Zhou}, \bibinfo{person}{Tian Fang}, {and} \bibinfo{person}{Long Quan}.} \bibinfo{year}{2020}\natexlab{}.
\newblock \showarticletitle{BlendedMVS: A Large-Scale Dataset for Generalized Multi-View Stereo Networks}. In \bibinfo{booktitle}{\emph{2020 IEEE/CVF Conference on Computer Vision and Pattern Recognition (CVPR)}}. \bibinfo{pages}{1787--1796}.
\newblock
\urldef\tempurl%
\url{https://doi.org/10.1109/CVPR42600.2020.00186}
\showDOI{\tempurl}


\bibitem[Yariv et~al\mbox{.}(2023)]%
        {baskedsdf}
\bibfield{author}{\bibinfo{person}{Lior Yariv}, \bibinfo{person}{Peter Hedman}, \bibinfo{person}{Christian Reiser}, \bibinfo{person}{Dor Verbin}, \bibinfo{person}{Pratul~P. Srinivasan}, \bibinfo{person}{Richard Szeliski}, \bibinfo{person}{Jonathan~T. Barron}, {and} \bibinfo{person}{Ben Mildenhall}.} \bibinfo{year}{2023}\natexlab{}.
\newblock \showarticletitle{BakedSDF: Meshing Neural SDFs for Real-Time View Synthesis}. In \bibinfo{booktitle}{\emph{ACM SIGGRAPH 2023 Conference Proceedings}} (Los Angeles, CA, USA) \emph{(\bibinfo{series}{SIGGRAPH '23})}. \bibinfo{publisher}{Association for Computing Machinery}, \bibinfo{address}{New York, NY, USA}, Article \bibinfo{articleno}{46}, \bibinfo{numpages}{9}~pages.
\newblock
\showISBNx{9798400701597}
\urldef\tempurl%
\url{https://doi.org/10.1145/3588432.3591536}
\showDOI{\tempurl}


\bibitem[Yu et~al\mbox{.}(2004)]%
        {yu2004mesh}
\bibfield{author}{\bibinfo{person}{Yizhou Yu}, \bibinfo{person}{Kun Zhou}, \bibinfo{person}{Dong Xu}, \bibinfo{person}{Xiaohan Shi}, \bibinfo{person}{Hujun Bao}, \bibinfo{person}{Baining Guo}, {and} \bibinfo{person}{Heung-Yeung Shum}.} \bibinfo{year}{2004}\natexlab{}.
\newblock \showarticletitle{Mesh Editing with Poisson-Based Gradient Field Manipulation}. In \bibinfo{booktitle}{\emph{ACM SIGGRAPH 2004 Papers}} (Los Angeles, California) \emph{(\bibinfo{series}{SIGGRAPH '04})}. \bibinfo{publisher}{Association for Computing Machinery}, \bibinfo{address}{New York, NY, USA}, \bibinfo{pages}{644–651}.
\newblock
\showISBNx{9781450378239}
\urldef\tempurl%
\url{https://doi.org/10.1145/1186562.1015774}
\showDOI{\tempurl}


\bibitem[Zhang et~al\mbox{.}(2021)]%
        {zhang2021editablefvv}
\bibfield{author}{\bibinfo{person}{Jiakai Zhang}, \bibinfo{person}{Xinhang Liu}, \bibinfo{person}{Xinyi Ye}, \bibinfo{person}{Fuqiang Zhao}, \bibinfo{person}{Yanshun Zhang}, \bibinfo{person}{Minye Wu}, \bibinfo{person}{Yingliang Zhang}, \bibinfo{person}{Lan Xu}, {and} \bibinfo{person}{Jingyi Yu}.} \bibinfo{year}{2021}\natexlab{}.
\newblock \showarticletitle{Editable Free-Viewpoint Video Using a Layered Neural Representation}.
\newblock \bibinfo{journal}{\emph{ACM Trans. Graph.}} \bibinfo{volume}{40}, \bibinfo{number}{4}, Article \bibinfo{articleno}{149} (\bibinfo{date}{jul} \bibinfo{year}{2021}), \bibinfo{numpages}{18}~pages.
\newblock
\showISSN{0730-0301}
\urldef\tempurl%
\url{https://doi.org/10.1145/3450626.3459756}
\showDOI{\tempurl}


\bibitem[Zhang et~al\mbox{.}(2023)]%
        {Zhang_2023_CVPR}
\bibfield{author}{\bibinfo{person}{Yunzhi Zhang}, \bibinfo{person}{Shangzhe Wu}, \bibinfo{person}{Noah Snavely}, {and} \bibinfo{person}{Jiajun Wu}.} \bibinfo{year}{2023}\natexlab{}.
\newblock \showarticletitle{Seeing a Rose in Five Thousand Ways}. In \bibinfo{booktitle}{\emph{Proceedings of the IEEE/CVF Conference on Computer Vision and Pattern Recognition (CVPR)}}. \bibinfo{pages}{962--971}.
\newblock


\bibitem[Zhou et~al\mbox{.}(2023)]%
        {zhou2023drivinggaussian}
\bibfield{author}{\bibinfo{person}{Xiaoyu Zhou}, \bibinfo{person}{Zhiwei Lin}, \bibinfo{person}{Xiaojun Shan}, \bibinfo{person}{Yongtao Wang}, \bibinfo{person}{Deqing Sun}, {and} \bibinfo{person}{Ming-Hsuan Yang}.} \bibinfo{year}{2023}\natexlab{}.
\newblock \bibinfo{title}{DrivingGaussian: Composite Gaussian Splatting for Surrounding Dynamic Autonomous Driving Scenes}.
\newblock
\newblock
\showeprint[arxiv]{2312.07920}~[cs.CV]


\end{thebibliography}

\appendix
\begin{figure*}
    \centering
    \includegraphics[width=\linewidth]{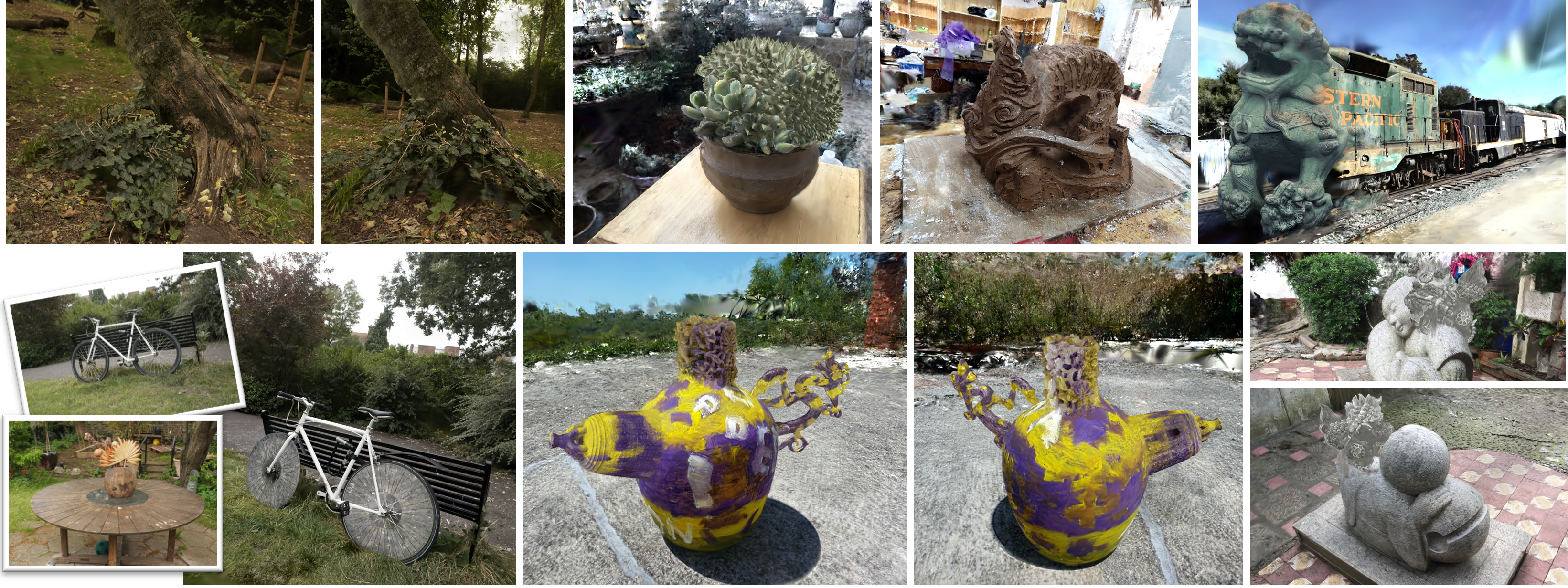}  
    \caption{Showcase in 3D with background. To demonstrate their natural appearance, we insert these composite models back into their unbounded backgrounds (the floaters are caused by the problem of 3DGS under unbounded scenes).
    %Showcase in 3D with background. We present these composite models by inserting them back into their unbounded backgrounds to illustrate their natural appearance (the floaters are due to the problem of 3DGS under unbounded scenes).
    }
    \label{fig:figure_only_insert}
\end{figure*}

\begin{figure*}
    \centering
    \includegraphics[width=\linewidth]{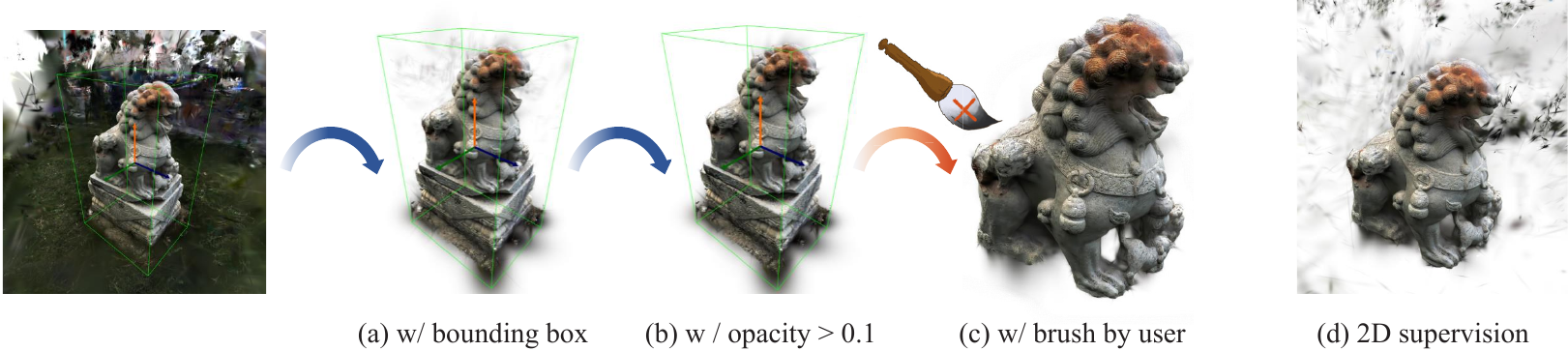} 
    \caption{We describe the segmentation workflow using our GUI and compare it to the result (d) from 2D mask supervision (for example, the Segment Anything Model (SAM) \cite{kirillov2023segany}). To segment with SAM, we re-implement the inverse-mask \cite{cen2023sa3d} strategy on 3DGS. A simple (a) bounding box with a (b) interactive (c) brush is demonstrated to be more practical in real-world scenes with numerous floaters. For more information, please refer to our supplementary video.} 
    %We describe the workflow for segmentation with our GUI and compare it to the result (d) from 2D mask supervision (e.g. segment anything model). As for segmenting with SAM, we re-implement the inverse-mask \cite{cen2023sa3d} strategy on 3DGS. A simple (a) bounding box with an (b) interactive (c) brush is shown to be more practical in real scenes with lots of floaters. See our supplementary video for more details.
    \label{fig:gui_segment}
\end{figure*}

\begin{figure*}
    \centering
    \includegraphics[width=\linewidth]{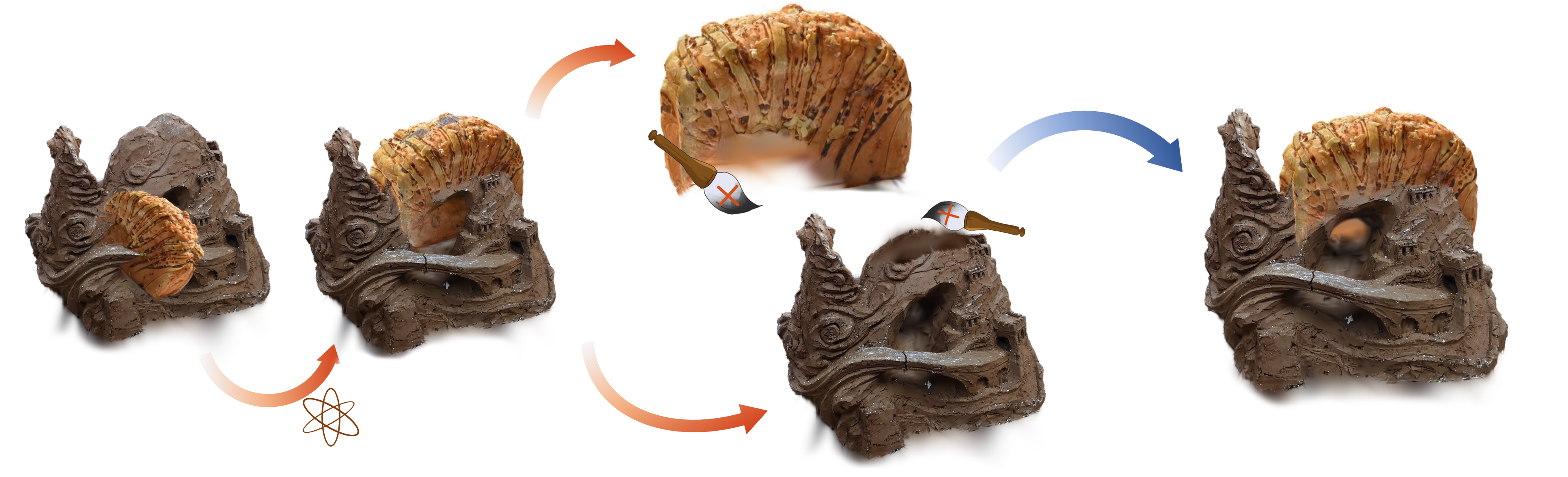} 
    \caption{We describe the transformation workflow using our GUI, as well as how to remove unwanted parts during composition. Users can adjust models to create a semantically meaningful composite, and then use a brush to remove unwanted parts, allowing for a more fine-grained composition. For more information, please refer to our supplementary video.}
    % We describe the workflow for transformation with our GUI and how to discard unwanted parts during composition. Users can first adjust models to obtain a semantically meaningful composite and then use a brush to discard unwanted parts, which supports a more fine-grained composition. See our supplementary video for more details.
    \label{fig:gui_transform}
\end{figure*}

\begin{figure*}
    \centering
    \begin{minipage}{\linewidth}
        \includegraphics[width=\linewidth]{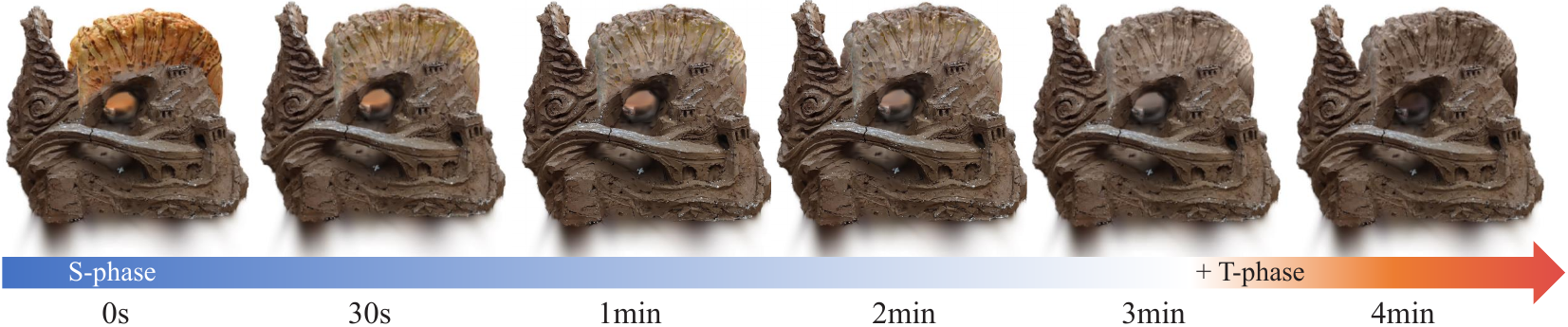}  
        \includegraphics[width=\linewidth]{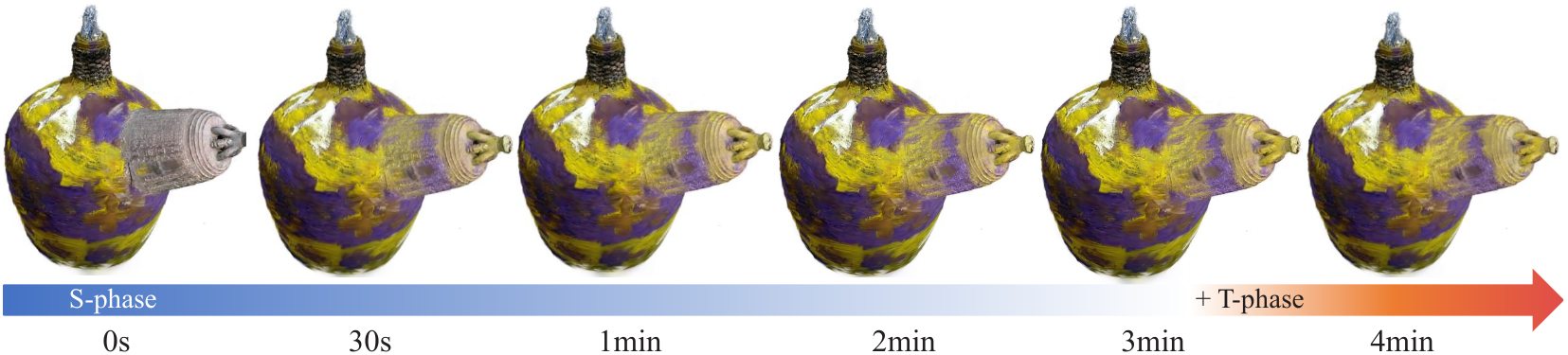}  
    \end{minipage}
    \caption{Visualize how our optimization gradually and efficiently converges. In our comparison, SNeRF \cite{meta2022snerf} takes over 10 hours, while SeamlessNeRF \cite{gong2023seamlessnerf} takes more than an hour. For more information, please refer to our supplementary video.}
    % Visualize how the optimization in ours gradually and efficiently converges. In contrast, SNeRF \cite{meta2022snerf} takes over 10 hours, and SeamlessNeRF \cite{gong2023seamlessnerf} takes over 1 hour in our comparison. See our supplementary video for more details.
    \label{fig:visualize_gradually_opt}
\end{figure*}

\begin{figure*}
    \centering
    \includegraphics[width=\linewidth]{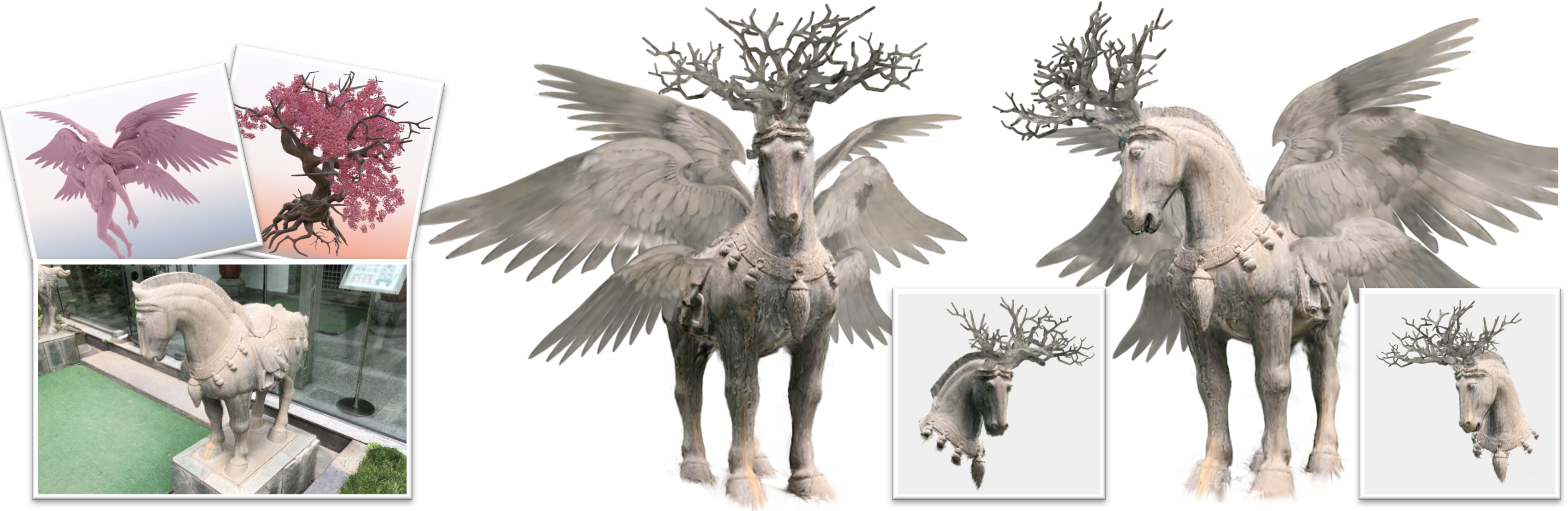}
    \caption{We demonstrate another compositing result using data from both the real world and the graphics engine. This additional case demonstrates our approach's versatility in dealing with both real and computer-generated models, validating its practical applicability. The two CG models were obtained from websites and rendered in Blender 3D on our own.}
%We show another compositing result where data are from both the real world and the graphics engine. This additional case serves to demonstrate the versatility of our approach in handling a combination of real and computer-generated models, further affirming its practical applicability. The two CG models were sourced from websites and rendered in Blender 3D on our own.}
    \label{fig:figure_only_cgmodel}
\end{figure*}

\end{document}